# Pulse Shape Discrimination Algorithms: Survey and Benchmark

Haoran Liu, Yihan Zhan, Mingzhe Liu, Yanhua Liu, Peng Li, Zhuo Zuo, Bingqi Liu, and Runxi Liu

*Abstract*— This review presents a comprehensive survey and benchmark of pulse shape discrimination (PSD) algorithms for radiation detection, classifying nearly sixty methods into statistical (time-domain, frequency-domain, neural network-based) and prior-knowledge (machine learning, deep learning) paradigms. We implement and evaluate all algorithms on two standardized datasets: an unlabeled set from a $^{241}$Am-$^9$Be source and a time-of-flight labeled set from a $^{238}$Pu-$^9$Be source, using metrics including Figure of Merit (FOM), F1-score, ROC-AUC, and inter-method correlations. Our analysis reveals that deep learning models, particularly Multi-Layer Perceptrons (MLPs) and hybrid approaches combining statistical features with neural regression, often outperform traditional methods. We discuss architectural suitabilities, the limitations of FOM, alternative evaluation metrics, and performance across energy thresholds. Accompanying this work, we release an open-source toolbox in Python and MATLAB, along with the datasets, to promote reproducibility and advance PSD research.

## I. INTRODUCTION

Pulse Shape Discrimination (PSD) is a critical technique in radiation detection [1], enabling the identification of different types of incident particles within a mixed radiation field. The utility of PSD is particularly evident in applications where desired signals must be isolated from an ever-present background of cosmic, terrestrial, or artificial radiation [2-8]. The technique's efficacy stems from the principle that different particles, upon interacting with a detector medium, generate electrical pulses with distinct temporal characteristics. By analyzing the shape of these pulses, it is possible to distinguish one type of particle from another, thereby enhancing measurement accuracy.

The physical basis for PSD is most prominently illustrated in the common task of separating neutrons and gamma-rays using organic scintillators. While both particles produce pulses with similar rising edges, their decay characteristics differ significantly. Gamma-rays, interacting primarily through Compton scattering with electrons, induce excitations to singlet states within the scintillator's molecules, which de-excite rapidly through prompt fluorescence. In contrast, neutrons scatter elastically from protons, and the resulting recoil protons create a high-density ionization track. This track preferentially populates molecular triplet states, which de-excite via slower mechanisms of delayed fluorescence and phosphorescence. Consequently, neutron-induced pulses exhibit a prolonged tail compared to the sharp, fast-decaying pulses from gamma-rays. This fundamental difference is exploited in fields requiring robust particle identification, such as nuclear reactor monitoring, radiopharmaceuticals, and fundamental physics research.

Over the decades, a diverse spectrum of algorithms has been developed to quantify and exploit these pulse shape differences. These methodologies can be broadly classified into two main paradigms: statistical discrimination and prior-knowledge discrimination [9]. The statistical discrimination approach operates on a corpus of collected pulses, for which a specific discrimination factor is calculated for each event. A histogram of these factors typically reveals distinct statistical distributions, often Gaussian in nature, corresponding to the different particle types. This category encompasses a wide range of feature extraction techniques. Seminal approaches operate in the time-domain, including methods such as Charge Comparison [10], Gatti's Parameter [11], and Pulse Gradient Analysis [12]. Later developments explored the frequency-domain, with algorithms like Frequency Gradient Analysis [13], Discrete Fourier Transform [14], and Wavelet Transform [15]. More recently, intelligent methodologies, classified here as neural network methods, have been applied within this statistical framework, including the Pulse-Coupled Neural Network [16], Ladder Gradient [17], and Spiking Cortical Model [18]. The prior-knowledge discrimination paradigm offers a different approach. Instead of relying on the statistical properties of a large dataset during analysis, these methods use a pre-labeled training dataset to build a predictive model. This model learns the salient features that distinguish between pulse

Manuscript received xx xxxxx 202x; revised xx xxxxx 202x; accepted xx xxxxx 202x. Date of publication xx xxxxx 202x; date of current version xx xxxxx 202x. This work was supported in part by the National Natural Science Foundation of China under Grant 12205078, and Grant 42104174; and in part by the Natural Science Foundation of Zhejiang Province under Grant LZ25F010007; and in part by the Wenzhou Major Science and Technology Innovation Project under Grant ZG2023011. (Corresponding author: Mingzhe Liu)

Haoran Liu and Mingzhe Liu are with the College of Nuclear Technology and Automation Engineering, Chengdu University of Technology, Chengdu 610059, China, and also with the School of Data Science and Artificial Intelligence, Wenzhou University of Technology, Wenzhou 325000, China (e-mail: liuhaoran@cdut.edu.cn, liumz@cdut.edu.cn)

Yihan Zhan, Peng Li, and Bingqi Liu are with the College of Nuclear Technology and Automation Engineering, Chengdu University of Technology, Chengdu 610059, China, (e-mail: zhanyihan@stu.cdut.edu.cn, lipeng@stu.cdut.edu.cn, liubingqi@cdu.edu.cn)

Yanhua Liu is with the College of Mechanical and Electrical Engineering, Chengdu University of Technology, Chengdu 610059, China (e-mail: liuyanhua@cdut.edu.cn).

Zhuo Zuo is with the College of Nuclear Technology and Automation Engineering, Chengdu University of Technology, Chengdu 610059, China, and also with the Southwestern Institute of Physics, Chengdu 610225, China (e-mail: zuozhuo@stu.cdut.edu.cn).

Runxi Liu is with the Faculty of Engineering Sciences, University College London, London WC1E 6BT, United Kingdom (e-mail: runxiliu@foxmail.com).

Color versions of one or more of the figures in this paper are available online at http://

Digital Object Identifier

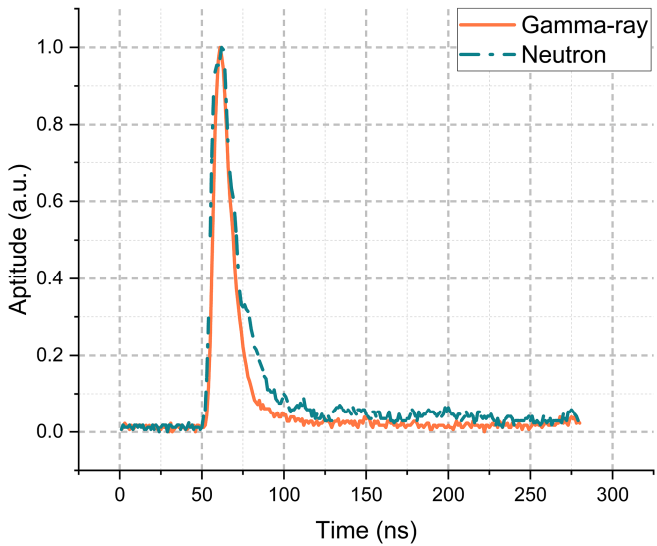

Figure 1. Example pulse signals

shapes and can subsequently classify individual incoming pulses, often in real-time. Within this paradigm, one can distinguish between classical machine learning methods, such as the Support Vector Machine [19], K-Nearest Neighbors [20], and Tempotron [9], and more recent deep learning models, including Multi-Layer Perceptrons [21], Convolutional Neural Networks [22], and the Transformer [23].

The field of PSD has been periodically summarized by several key review articles, each providing a valuable but distinct perspective on its progress. Much of the recent focus has been on advancements in detector materials. For instance, reviews by Bertrand et al. (2015) [24] and Zaitseva et al. (2018) [25] provided in-depth surveys from chemistry and materials science standpoints, detailing the synthesis, composition, and physical properties of novel plastic scintillators with enhanced PSD capabilities. Other works have concentrated on specific technological breakthroughs, such as the review by Foster et al. (2020) [26] on heterogeneous composite detectors for achieving triple-pulse-shape-discrimination. More recently, the review by Ahnouz et al. (2024) [27] shifted the focus back to signal processing, offering a broad survey comparing classical PSD algorithms against modern machine learning approaches.

While these articles provide essential overviews of material advancements, specific technologies, and high-level algorithmic surveys, a critical gap remains in the literature. The sheer volume and diversity of proposed algorithms, often evaluated on disparate datasets with varying metrics, make it exceedingly difficult for researchers and engineers to perform direct, fair comparisons and select the optimal method for their specific needs. No existing work has undertaken a large-scale, systematic, and quantitative benchmark of these numerous techniques on a unified experimental and computational platform. This paper aims to fill that void. We present the most comprehensive comparative study of PSD algorithms to date, implementing and evaluating nearly sixty distinct methods from across the literature. To ensure a rigorous and impartial assessment, all algorithms were benchmarked on two standardized datasets which we have made publicly available in conjunction with this study. Furthermore, a principal contribution of this work is the release of a comprehensive open-source library containing the implementations of all surveyed algorithms in both MATLAB and Python, in order to satisfy the needs of various research communities. This dual-language resource is designed to promote reproducibility and serve as an enduring tool for researchers. This review, therefore, not only surveys the vast landscape of PSD but also provides the critical, data-driven insights necessary to navigate it effectively.

The remainder of this review is organized as follows. We begin by systematically presenting the theoretical foundations and operational principles of the surveyed PSD algorithms, structured according to our proposed taxonomy. The review first addresses the statistical discrimination paradigm, with Section II dedicated to time-domain methods, Section III covering frequency-domain techniques, and Section IV examining neural network-based approaches. Subsequently, we explore the prior-knowledge paradigm, detailing classical machine learning methods in Section V and modern deep learning architectures in Section VI. With the fundamentals of all methods established, Section VII presents our comprehensive comparative study, outlining the experimental framework and reporting the benchmark results. Finally, Section VIII provides a discussion of the key findings and their broader implications, followed by Section IX, which concludes the paper with a summary of our contributions and perspectives on future research directions.

## II. TIME-DOMAIN PSD ALGORITHM

Time-domain methods for PSD focus on analyzing the temporal characteristics of scintillation pulses directly as they are captured by a digitizer. These techniques extract features from the raw waveform by examining its shape, timing, and amplitude properties over time. The fundamental principle behind these methods is that different types of incident particles (e.g., neutrons and gamma-rays) produce pulses with distinct shapes, particularly in their decay times, as shown in Figure 1.

### A. Charge Comparison

The Charge Comparison (CC) method [10] is a classic PSD technique that differentiates particles based on the distribution of charge in their scintillation pulses. The core idea is to compare the charge collected in the tail (slow component) of the pulse to the total charge of the pulse. Neutron-induced pulses typically have a more significant slow component compared to gamma-ray-induced pulses. This method calculates a discrimination factor by taking the ratio of the slow component's charge to the total charge:

$$F = \frac{\int_{t_{peak}+t_{short}}^{t_{peak}+t_{long}} V(t)}{\int_{t_{peak}-t_{pre}}^{t_{peak}+t_{long}} V(t)}. \quad (1)$$

where $V(t)$ is the pulse amplitude at time $t$, $t_{peak}$ is the time of the pulse peak, and $t_{pre}$, $t_{short}$, $t_{long}$ are the gate parameters.

### B. Charge Integration

The Charge Integration (CI) method [28], similar to the CC method, relies on comparing charge over different time intervals. It calculates the discrimination factor by taking the ratio of the charge in a delayed gate to the total charge of the pulse:

$$F = \frac{\int_{t_{peak}+t_{total}-t_{delay}}^{t_{peak}+t_{total}} V(t)}{\int_{t_{peak}-t_{pre}}^{t_{peak}+t_{total}} V(t)}. \quad (2)$$

where $t_{pre}$, $t_{delay}$, $t_{total}$ are the gate parameters. Indeed, the CI and CC methods are distinguished by their original hardware circuit designs, which use different techniques to collect charge. In a digital environment, this difference is abstracted away, and the two methods become fundamentally equivalent, both relying on numerical integration of the pulse waveform.

*C. Falling-Edge Percentage Slope*

The Falling-Edge Percentage Slope (FEPS) method [29] provides discrimination by analyzing the slope of the pulse's falling edge. It calculates the slope between two points on the trailing edge that correspond to fixed percentages (e.g., 60% and 10%) of the pulse's maximum amplitude:

$$F = \frac{V(t_{10\%})-V(t_{60\%})}{t_{10\%}-t_{60\%}}. \quad (3)$$

where $t_{10\%}$ and $t_{60\%}$ are the times at which the pulse amplitude on the falling edge is 10% and 60% of the maximum amplitude, respectively. Since neutron pulses have a slower decay time, their falling-edge slope is typically less steep than that of gamma-ray pulses.

*D. Gatti Parameter*

The Gatti Parameter (GP) method [11] is a weighted linear technique that uses reference signals for both particle classes (e.g., neutrons and gammas) to create an optimal weighting function. The discrimination factor is computed as a weighted sum of the pulse samples. The weight function, known as the Gatti Parameter, is derived from the reference signals to maximize the separation between the two classes:

$$P(t) = \frac{V_n(t)-V_\gamma(t)}{V_n(t)+V_\gamma(t)}. \quad (4)$$

where $V_n(t)$ and $V_\gamma(t)$ are the normalized reference pulses for neutrons and gamma-rays, respectively. The PSD factor $F$ is then:

$$F = \sum_t V(t) \cdot P(t). \quad (5)$$

*E. Log-Likelihood Ratio*

The Log-Likelihood Ratio (LLR) method [30, 31] is a probabilistic approach that calculates a discrimination factor based on the probability mass function (PMF) of the pulse shapes. It compares the likelihood that a given pulse belongs to one particle class versus another. The PMF is calculated from reference signals for each class, which is then used for PSD factor $F$ calculations:

$$F = \sum_t V(t) \cdot (-\log(\frac{PMF_n(t)}{PMF_\gamma(t)})). \quad (6)$$

*F. Log Mean Time*

The Log Mean Time (LMT) method [32] calculates the amplitude-weighted mean time of the pulse and then takes its natural logarithm to produce a discrimination factor:

$$F = \log(\frac{\sum_t t \cdot V(t)}{\sum_t V(t)}). \quad (7)$$

The mean time represents the temporal center of the pulse, which is influenced by the decay time. Neutron pulses, with their longer tails, will have a larger mean time compared to gamma pulses.

*G. Principal Component Analysis*

The Principal Component Analysis (PCA) method [33] is a dimensionality reduction technique applied to PSD. It identifies the direction of maximum variance (the first principal component) in a training set of pulses. The PSD factor $F$ is the projection of the pulse signal onto the first principal component:

$$F = |V(t) \cdot W_{PC1}|. \quad (8)$$

where $W_{PC1}$ is the first principal component eigenvector of the covariance matrix of a training set.

*H. Pulse Gradient Analysis*

The Pulse Gradient Analysis (PGA) method [12] is a straightforward technique that calculates the gradient of the pulse waveform at a specific point. It measures the rate of change of the pulse amplitude between the peak and a fixed time point after the peak:

$$F = \frac{V(t_{peak}+\Delta t)-V(t_{peak})}{\Delta t}. \quad (9)$$

where $\Delta t$ is a fixed time interval after the peak. The PGA method is conceptually similar to FEPS, as both techniques calculate the gradient of the pulse's tail. The primary distinction lies in how the interval for the gradient calculation is determined: PGA uses a fixed time interval relative to the pulse peak, whereas FEPS defines the interval based on points corresponding to specific percentages of the peak amplitude.

*I. Pattern Recognition*

The Pattern Recognition (PR) method [34] treats pulses as vectors in a high-dimensional space and uses the angle between them as a measure of similarity. A reference pulse should be chosen first, the PSD factor $F$ is then the angle between the pulse vector $V$ and the reference vector $V_{ref}$:

$$F = \arccos\left(\frac{V \cdot V_{ref}}{|V||V_{ref}|}\right). \quad (10)$$

*J. Zero-Crossing*

The Zero-Crossing (ZC) method [35, 36] involves transforming the original unipolar pulse into a bipolar pulse using a recursive filter. The discrimination factor is then determined by the time at which the bipolar pulse crosses the zero-voltage level:

$$F = t_{zero\_crossing} - t_{start}. \quad (11)$$

where $t_{zero\_crossing}$ is the time of zero-crossing point of the filtered bipolar pulse and $t_{start}$ is a reference start time, often a fraction of the time to the peak of the pulse.

A key aspect of the Zero Crossing method is the application of a third-order recursive filter, which transforms the original unipolar pulse into a bipolar signal. This transformation is engineered to produce a curve whose characteristics approximate the second derivative of the input signal.

Consequently, the point where the filtered signal crosses the zero-axis corresponds directly to the inflection point on the tail of the original pulse—the moment of its most rapid descent. As a result, Neutron pulses, with their slower decay time, will have a larger zero-crossing time compared to gamma pulses.

## III. FREQUENCY-DOMAIN PSD ALGORITHM

Frequency-domain methods for PSD transform the time-domain pulse signals into the frequency domain to analyze their spectral characteristics. These methods are based on the principle that the different temporal shapes of pulses from various particle types (e.g., neutrons and gamma-rays) translate into distinguishable features in their frequency spectra.

### A. Discrete Fourier Transform

The Discrete Fourier Transform (DFT) method [14] analyzes the zero-frequency components of the discrete cosine and sine transforms (DCT and DST) of the pulse signal. The discrimination factor is computed as a ratio of the sum of the squared DFT components to the product of the zero-frequency components of the DST and DCT, normalized by the sum of the trimmed signal:

$$F = \frac{\sum_t |DFT(V_{trim}(t))|^2}{DST(V_{trim}(t))_0 \cdot DCT(V_{trim}(t))_0} \cdot \frac{1}{\sum_t V_{trim}(t)}. \quad (12)$$

where $V_{trim}(t)$ is the pulse signal from its peak onward.

### B. Frequency Gradient Analysis

The Frequency Gradient Analysis (FGA) method [13] computes the gradient between the first two frequency components of the signal using the Fourier transform:

$$F = L \cdot \frac{|X_0 - X_1|}{f_s}. \quad (13)$$

$$X_0 = \sum_t V(t). \quad (14)$$

$$X_1 = \left|\sum_t \left(V(t) \cos\left(\frac{2\pi t}{L}\right)\right)\right| - \sum_t \left(V(t) \sin\left(\frac{2\pi t}{L}\right)\right). \quad (15)$$

where $L$ is the signal length, $f_s$ is the sample frequency, $X_0$ and $X_1$ are the zeroth and first components of the Fourier spectrum.

### C. Fractal Spectrum

The Fractal Spectrum (FS) method [37] analyzes the fractal dimension of the pulse signals in the frequency domain. First, the power spectrum of the pulse is calculated using a periodogram. The method then involves a log-log transformation of both the power spectrum and its corresponding frequencies. By performing a linear regression on this transformed data, the slope is determined and used to compute a final discrimination factor:

$$F = \frac{b}{a} - slope. \quad (16)$$

where $a$ and $b$ are empirical constants.

### D. Scalogram-based Discrimination

The Scalogram-based Discrimination (SD) method [38] transforms one-dimensional time-domain pulses into two-dimensional time-frequency representations to facilitate particle identification. At its core, the method employs the Continuous Wavelet Transform (CWT) to generate a matrix of coefficients for each pulse, representing its energy distribution across time and frequency scales. This scalogram is subsequently binarized by applying an intensity threshold to produce a binarized scalogram, $B(t,f)$, which highlights the signal's most energetic components.

To distinguish between particle types, the SD method utilizes labeled reference signals to define a specific discrimination mask, $M(t,f)$. This mask is derived by systematically comparing the binarized scalograms of different signal classes and identifying the time-frequency coordinates where their structural differences are most pronounced and consistent. When classifying an unlabeled signal, this pre-defined mask is applied to its binarized scalogram, and a PSD factor $F$ is computed by summing the pixel values within the masked area:

$$F = \frac{\sum_{t,f} B(t,f) \cdot M(t,f)}{\sum_{t,f} M(t,f)}. \quad (17)$$

where $t$ and $f$ represents the time and frequency scales, respectively, and $\sum_{t,f} M(t,f)$ is a normalization factor corresponding to the total area (i.e., the number of non-zero pixels) of the discrimination mask.

### E. Simplified Digital Charge Collection

The Simplified Digital Charge Collection (SDCC) method [39] is a simplified domain transformation approach that analyzes the decay rate differences in the frequency domain. The discrimination factor is computed as the natural logarithm of the sum of the squared values of the pulse signal:

$$F = \log(\sum_t V(t)^2). \quad (18)$$

### F. Wavelet Transform (WT1 - Haar)

This Wavelet Transform (WT1) method [15] uses the Haar wavelet to transform the pulse signal. The discrimination factor is computed as the ratio of specific scale features (e.g., the 28th and 40th) from the resulting scale function, which are chosen to maximize the separation between neutron and gamma signals. The PSD factor $F$ is calculated as:

$$F = \sqrt{\frac{s_2}{s_1} \frac{\int |t \cdot (V * \psi_{s_1}^{Haar})(t)| dt}{\int |t \cdot (V * \psi_{s_2}^{Haar})(t)| dt}}. \quad (19)$$

where $(V * \psi_s^{Haar})(t)$ is the convolution between the signal $V(t)$ and the scaled Haar Wavelet $\psi_s^{Haar}(t)$ (i.e., the Haar Wavelet transform of the pulse signal at scale $s$).

### G. Wavelet Transform (WT2 - Marr)

This Wavelet Transform (WT2) method [40] uses the Marr (Mexican Hat) wavelet. The discrimination factor is computed as a ratio of the integral of the original signal to the sum of the integral of the original signal and the integral of the positive part of the wavelet-transformed signal:

$$F = \frac{2 \int V(t) dt}{\int V(t) dt + \int \max(0, (V * \psi^{Marr})(t)) dt}. \quad (20)$$

where $(V * \psi^{Marr})(t)$ is the convolution between the signal $V(t)$ and the Marr Wavelet $\psi^{Marr}(t)$ (i.e., the Marr Wavelet transform of the pulse signal).

## IV. NEURAL NETWORK PSD ALGORITHM

Neural network-based methods for PSD leverage the powerful feature extraction and pattern recognition capabilities of artificial neural networks. These methods, particularly those inspired by biological neural systems like spiking neural networks, can analyze temporal patterns and extract discrimination features directly from raw or minimally processed pulse waveforms, without pre-training process. Instead of relying on hand-crafted features, these models use layers of interconnected neurons to automatically identify the most salient characteristics that distinguish between different particle types.

*A. Ladder Gradient*

The Ladder Gradient (LG) method [17] utilizes a Quasi-Continuous Spiking Cortical Model (QCSCM) to transform the input pulse into an ignition map. This map, denoted $I^{QCSCM}(t)$, is generated by simulating a network of spiking neurons whose firing activity is driven by the input signal. The value $I^{QCSCM}(t)$ at each time point represents the cumulative number of neuron firings, effectively encoding the dynamic characteristics of the pulse shape. The final discrimination factor is then calculated as the slope between two key points on this map:

$$F = \frac{I^{QCSCM}(t_{max}) - I^{QCSCM}(t_{mode})}{t_{max} - t_{mode}}. \quad (21)$$

where $t_{max}$ is the time index corresponding to the peak of the original input signal, and $t_{mode}$ is the $m$-th element of the set $T_{mode}$:

$$T_{mode} = \{t \mid t > t_{max} \text{ and } I^{QCSCM}(t) = mode\{I^{QCSCM}(t)\}\}. \quad (22)$$

where $mode\{I^{QCSCM}(t)\}$ denotes the most frequent value (the mode) in the QCSCM ignition map.

*B. Pulse-Coupled Neural Network*

The Pulse-Coupled Neural Network (PCNN) method [16, 41] first transforms the input pulse signal, $V(t)$, into an ignition map $I^{PCNN}(t)$, similar to the LG method. The PCNN serves as the prototype model for a family of spiking neural networks used in pulse shape discrimination, including the QC-SCM (used in the LG method) and several other models described in this neural network PSD algorithm section. It is a biologically inspired model that emulates the behavior of neurons in the cat's visual cortex. The PSD factor $F$ is simply computed as the sum of all firings for each neuron over time:

$$F = \sum_t I^{PCNN}(t). \quad (23)$$

*C. Random-Coupled Neural Network*

The Random-Coupled Neural Network (RCNN) method [42] utilizes another variant in the family of PCNN-based models, RCNN. Its primary innovation is the introduction of a dynamic, randomized coupling mechanism, which distinguishes it from the deterministic nature of the standard PCNN. While the PCNN uses a single, fixed connection weight matrix for the entire simulation, the RCNN generates a new, random weight matrix at every single iteration. This introduction of a stochastic element enhances the network's sensitivity to subtle pulse shape variations, resulting in ignition maps with features that are more robust and effective for PSD. The PSD factor F is also computed as the sum of all firings for each neuron over time:

$$F = \sum_t I^{RCNN}(t). \quad (24)$$

*D. Spiking Cortical Model*

The Spiking Cortical Model (SCM) method [18] is based on another variant in the family of PCNN-like models, SCM. It simplifies the full PCNN structure by removing the separate linking and feedback pathways, combining them into a single internal state update. Although structurally similar to the QC-SCM from the Ladder Gradient method, the SCM differs in key implementation details. The primary distinction lies in the simulation's time step: SCM uses discrete, integer steps, whereas QC-SCM uses fractional steps to better approximate a continuous-time system. Additionally, instead of the gradient-based PSD factor calculation of LG, SCM sums all firings for each neuron over the total number of iterations:

$$F = \sum_t I^{SCM}(t). \quad (25)$$

## V. DEEP LEARNING PSD ALGORITHM

Deep learning (DL) methods represent a paradigm shift in PSD, moving from statistical feature analysis to data-driven feature learning. These methods employ neural networks with various architectures to learn hierarchical representations of the pulse data, enabling them to capture highly complex and subtle patterns that may be missed by traditional statistical techniques. By training on large datasets of labeled pulses, these models can learn an end-to-end mapping from the raw pulse waveform (or a transformed representation) to a desired output. This output can be either a direct classification of the particle type or a continuous PSD factor, which allows for subsequent statistical analysis. Reflecting this versatility, all DL methods investigated in this work are systematically evaluated for both classification and regression tasks to provide a comprehensive performance analysis. Limited by the scope of this review, we do not cover the fundamentals of DL. Readers unfamiliar with this topic are referred to [43] for further details.

*A. Convolutional Neural Networks*

Convolutional Neural Networks (CNNs) are a class of DL models particularly adept at processing grid-like data [44]. Their architecture is inspired by the human visual cortex, employing convolutional layers with learnable filters (or kernels) to automatically and hierarchically detect patterns. For PSD, both 1D and 2D CNNs are utilized. 1D CNNs can be applied directly to the raw pulse waveforms to learn temporal features, while 2D CNNs are typically used to process time-frequency representations of the signal, such as spectrograms or scalograms, or even 2D images of the pulse shape itself. By stacking layers, CNNs can build up a rich representation, from simple local edges in early layers to complex, abstract features in deeper layers. To provide a comprehensive overview, this study evaluates several distinct CNN architectures, varying in depth, input dimensionality, and feature representation.

*1D-CNN (CNNShallow):* This model is a 1D CNN applied directly to the raw pulse waveforms [22]. Its architecture consists of two convolutional layers with max pooling, followed by an adaptive average pooling layer and an output layer. It is designed to learn discriminative features directly from the time-domain signal with a relatively simple and computationally efficient architecture.

*2D-CNN (CNNDeep):* In contrast to the shallow model, this is a much deeper 2D CNN architecture. It takes as input a time-frequency representation of the pulse, specifically a scalogram generated using the CWT. The network contains four blocks of convolutional layers, progressively increasing in channel depth, allowing it to learn highly complex and hierarchical features from the 2D scalogram images.

*CNN with Snapshot Plotting (CNNSP):* This unique approach converts each 1D pulse waveform into a 2D snapshot image by plotting its shape [45]. A moderately deep 2D CNN, with three convolutional layers, is then trained to perform image classification or regression on these pulse snapshots. This method tests the ability of a standard image-based CNN to distinguish the graphical representations of neutron and gamma pulses.

*CNN with Fourier Spectrogram Features (CNNFT):* This model uses a 2D CNN to process a power spectrogram of the pulse signal, which is efficiently computed using the Fast Fourier Transform (FFT). The network architecture consists of three convolutional layers, designed to learn patterns from the resulting time-frequency representation.

*CNN with STFT Features (CNNSTFT):* This variant uses a similar three-layer 2D CNN architecture. However, instead of a power spectrogram, its input is a magnitude spectrogram computed directly from the Short-Time Fourier Transform (STFT). This provides a slightly different representation of the time-frequency information for the network to learn from.

*CNN with Wavelet Scalogram Features (CNNWT):* This model uses a 2D CNN architecture with three convolutional layers, similar to the STFT-based models. However, its input is a scalogram generated using CWT. This allows the network to leverage the multi-resolution analysis capabilities of wavelets to potentially capture transient features in the pulse more effectively.

*B. Multi-Layer Perceptrons*

Multi-Layer Perceptrons (MLPs) are a foundational class of feedforward artificial neural networks [46] and often serve as a benchmark model in DL-based PSD. An MLP processes input data by passing it through a series of interconnected layers. For PSD, the input is typically a one-dimensional vector, which can be the raw, flattened pulse waveform or a set of hand-crafted features extracted through other signal processing techniques. The core architecture consists of an input layer, one or more hidden layers, and an output layer. Each layer contains a number of neurons, and each neuron in a hidden layer applies a non-linear activation function (such as the Rectified Linear Unit, ReLU) to its weighted input. This introduction of non-linearity is crucial, as it allows the network to learn complex, non-linear relationships between the input features and the output. The network learns by iteratively adjusting its weights and biases through a process called backpropagation, driven by an optimization algorithm that seeks to minimize a task-specific loss function.

MLPs are highly versatile and can be adapted for both classification and regression tasks. For classification, the goal is to directly assign a particle label (e.g., neutron or gamma-ray), and the output layer typically uses a sigmoid activation function for binary classification or a softmax function for multi-class scenarios to provide a probability distribution over the classes. For regression, the objective is to predict a continuous PSD factor, analogous to the output of traditional methods. In this case, the output layer usually consists of a single neuron with a linear activation function, and this continuous output can be used for statistical feature analysis.

To provide a comprehensive assessment of MLP capabilities for the PSD task, this study not only considers previously proposed architectures but also systematically explores a range of novel variants distinguished by their depth and input features.

*Single Layer Perceptron (MLP1):* This model consists of a single fully-connected layer mapping the raw flattened pulse waveform directly to a single output neuron with a sigmoid activation function. It contains no hidden layers and essentially performs logistic regression, serving as a simple linear baseline.

*Two-Hidden-Layer Perceptron (MLP2):* Based on the Dense Neural Network (DNN) from [47], this model first extracts six features from each pulse corresponding to the cumulative charge at different truncation points. These features are then fed into a network with two hidden layers (each with 10 neurons and dropout) before the output layer.

*Deep Multilayer Perceptron (MLP3):* This model applies a deep feedforward network with seven hidden layers directly to the raw pulse waveform [21]. It is designed to test the capabilities of a very deep architecture in learning complex patterns from the signal without manual feature engineering.

*Single-Layer Perceptron with Fourier Transform Features (MLP1FT):* In this hybrid model, a single fully-connected layer is applied not to the raw pulse but to the magnitude spectrum obtained from its FFT. This tests the effectiveness of a simple linear model on frequency-domain features.

*Two-Hidden-Layer Perceptron with Fourier Transform Features (MLP2FT):* This model combines a two-hidden-layer architecture with Fourier transform features. It investigates whether a deeper network can better exploit the frequency-domain information to improve PSD performance compared to its single-layer counterpart.

*Deep Multilayer Perceptron with Fourier Transform Features (MLP3FT):* This variant uses the seven-hidden-layer MLP architecture with the FFT magnitude spectrum as input, exploring the potential benefits of maximum network depth when operating on frequency-domain features.

*Single-Layer Perceptron with Principal Component Analysis Features (MLP1PCA):* This model uses a single fully-connected layer where the input is the first 100 principal components of the pulse, derived via PCA. This approach tests the effectiveness of a linear model on dimensionality-reduced features.

*Two-Hidden-Layer Perceptron with Principal Component Analysis Features (MLP2PCA):* This variant employs the deeper two-hidden-layer architecture on the principal component features of the pulse. It assesses whether additional network capacity can leverage the compressed information from PCA more effectively.

*Deep Multilayer Perceptron with Principal Component Analysis Features (MLP3PCA):* The deep seven-hidden-

layer architecture is applied to the PCA-reduced features. This model explores the performance when combining a powerful dimensionality reduction technique with a very deep feedforward network.

*Single-Layer Perceptron with Short-Time Fourier Transform Features (MLP1STFT):* This model feeds the flattened magnitude of the STFT of the pulse into a single fully-connected layer. This allows the network to learn from a time-frequency representation of the signal.

*Two-Hidden-Layer Perceptron with Short-Time Fourier Transform Features (MLP2STFT):* Combining the two-hidden-layer architecture with STFT features, this model investigates if a deeper network can better interpret the complex time-frequency patterns for more accurate discrimination.

*Deep Multilayer Perceptron with Short-Time Fourier Transform Features (MLP3STFT):* This variant uses the deep seven-hidden-layer architecture to process flattened STFT features, testing the limits of a feedforward network in learning from detailed time-frequency information.

*Single-Layer Perceptron with Wavelet Transform Features (MLP1WT):* In this model, the input to a single fully-connected layer is the concatenated approximation and detail coefficients from a single-level Discrete Wavelet Transform (DWT) of the pulse.

*Two-Hidden-Layer Perceptron with Wavelet Transform Features (MLP2WT):* This model uses the two-hidden-layer architecture on the DWT features. It aims to determine if a deeper network can take better advantage of the multi-scale information provided by the DWT.

*Deep Multilayer Perceptron with Wavelet Transform Features (MLP3WT):* The deep seven-hidden-layer architecture is combined with DWT features in this variant. It explores the synergy between a very deep network and a wavelet-based feature representation.

## C. Recurrent Neural Networks

Recurrent Neural Networks (RNNs) are a class of neural networks specifically designed for sequential data [48], making them naturally suited for analyzing time-series like pulse waveforms. Unlike feedforward networks, RNNs have connections that loop back on themselves, creating an internal state that allows them to capture temporal dependencies in the data. The network processes a pulse step-by-step, updating its hidden state at each point based on the current input and the previous state. This makes them powerful for learning the dynamic, time-varying features that distinguish particle types. This study evaluates several RNN architectures, including the simple RNN and more advanced gated variants developed to better handle long-range dependencies.

*Simple Recurrent Neural Network (RNN):* This model uses a standard RNN layer with 5 hidden units, which processes the input pulse sequence. The hidden state from the final time step is then passed through a two-layer feedforward network (15 neurons, then 1 output neuron) to produce the discrimination result. This serves as a baseline for recurrent architectures.

*Elman Neural Network (ENN):* Based on the architecture proposed in [49], this model first projects the raw input pulse into a high-dimensional space (80 features) using a fully-connected layer. The resulting sequence is then processed by a simple RNN layer (20 hidden units). The final hidden state is subsequently passed through a two-layer feedforward network (15 neurons, then 1 output neuron) for the final prediction.

*Long Short-Term Memory (LSTM):* This model employs an LSTM layer with 5 hidden units [47]. LSTMs are an advanced RNN variant designed to overcome the vanishing gradient problem and learn long-range dependencies using a gated memory cell. Similar to the other recurrent models, the final hidden state from the LSTM layer is fed into a two-layer feedforward network (15 neurons, then 1 output neuron) to produce the output.

*Gated Recurrent Unit (GRU):* This model utilizes a GRU layer, which is a more recent and slightly simpler gated variant of the LSTM. The architecture consists of a GRU layer with 5 hidden units, whose final hidden state is then processed by a two-layer feedforward network (15 neurons, then 1 output neuron) to generate the discrimination factor.

## D. Transformer

The Transformer architecture established a novel approach for sequence modeling by replacing conventional recurrent and convolutional operations with a self-attention mechanism [23]. Unlike RNNs that process data sequentially, the Transformer's design allows for parallel processing of all elements in a sequence. At its core, the self-attention layer enables the model to dynamically weigh the influence of different parts of the input when generating a representation for a specific position. To compensate for the absence of inherent sequential processing, positional encodings are incorporated into the input embeddings, providing the model with crucial information about the relative or absolute position of each data point in the sequence.

For this work, the Transformer architecture is adapted to process raw pulse waveforms. Each pulse is first projected into a 64-dimensional embedding space, to which positional encodings are added to preserve temporal information. The central component of the model is a two-layer stack of Transformer encoders, where each layer utilizes two attention heads. The final representation is obtained by applying mean pooling across the output sequence, which is then passed to a linear layer for prediction.

## E. Mamba

Mamba has recently emerged as a powerful and efficient alternative to Transformers for modeling long sequences [50]. It is founded on the principles of structured state-space models (SSMs), which are adapted from classical control theory to represent continuous-time systems. Mamba augments the standard SSM framework with an input-dependent selection mechanism. This allows the model to selectively focus on or filter out information from the input sequence, enabling it to dynamically modulate the flow of information. A primary advantage of this design is its linear-time complexity with respect to sequence length, making it substantially more computationally efficient for capturing long-range dependencies than the quadratic-complexity self-attention mechanism in standard Transformers.

In this study, the Mamba architecture is applied to raw pulse waveforms. The input signal is projected into a 16-dimensional space and subsequently processed by a stack of Mamba blocks—one block for the classification task and two for regression. The output from the final block is aggregated using mean pooling before being fed to a terminal linear layer for prediction.

## VI. Machine learning PSD algorithm

Machine learning (ML) methods provide another framework for PSD, often serving as a bridge between traditional statistical techniques and DL models. The first stage of these methods often involves a manual or semi-automated feature extraction process designed to transform the high-dimensional raw pulse waveform into a compact and informative low-dimensional feature space. The choice of feature extraction technique varies significantly across different models, reflecting diverse strategies to capture the essential discriminatory information within the pulse shape.

In the second stage, a ML algorithm is trained on these extracted features to perform either classification (distinguishing neutrons from gamma-rays) or regression (predicting a continuous PSD parameter). This modular approach allows for the combination of expert knowledge in feature design with the powerful pattern recognition capabilities of ML classifiers and regressors. Limited by the scope of this review, we do not cover the fundamentals of ML. Readers unfamiliar with this topic are referred to [51] for further details.

### A. Boosted Decision Tree

The Boosted Decision Tree (BDT) is an ensemble learning method that combines many weak decision trees to create a strong classifier or regressor. It uses a boosting algorithm, typically AdaBoost, which iteratively trains new trees on a reweighted version of the training data, giving more weight to instances that were previously misclassified. This allows the model to focus on difficult-to-classify examples and often leads to better performance than a single decision tree.

### B. Decision Tree

The Decision Tree (DT) is a non-parametric supervised learning method that learns simple decision rules inferred from the data features to predict the target variable. For PSD, a decision tree can be used to classify pulses by partitioning the feature space based on a series of learned thresholds [52]. To handle the high dimensionality of the pulse data, PCA is first used to reduce the data to a 2D feature space.

### C. Fuzzy C-Means

The Fuzzy C-Means (FCM) is an unsupervised clustering algorithm that allows each data point to belong to multiple clusters with varying degrees of membership. In the context of PSD, it can be used to cluster pulses into neutron and gamma-ray categories [53]. Unlike hard clustering algorithms, FCM provides a degree of membership for each pulse to each class, which can be useful for handling ambiguous or overlapping pulses.

### D. Gaussian Mixture Model

The Gaussian Mixture Model (GMM) is an unsupervised probabilistic model that assumes all the data points are generated from a mixture of a finite number of Gaussian distributions with unknown parameters. For PSD, a GMM can be used to model the distribution of neutron and gamma-ray pulses in a feature space [54]. The features used are the total charge of the pulse and the tail-to-total integral ratio.

### E. K-Nearest Neighbors

The K-Nearest Neighbors (KNN) is a supervised non-parametric, instance-based learning algorithm [20]. For classification, it assigns a new data point to the class that is most common among its k-nearest neighbors in the feature space. For regression, it predicts the average values of its k-nearest neighbors. The features are derived by dividing the pulse into a fixed number of segments and calculating the sum of each segment.

### F. Linear and Logistic Regression

Linear and Logistic Regression are fundamental supervised statistical models for regression and classification, respectively.

*Linear Regression (**LINRE**):* This model learns a linear relationship between the input features and a continuous output variable. For PSD, it can be used to predict a continuous PSD factor. The input features are the first two principal components of the pulse data.

*Logistic Regression (**LOGRE**):* This model is used for binary classification. It learns a linear decision boundary to separate the two classes. The input features are also the first two principal components.

*Logistic Regression with STFT (**LRSTFT**):* This is a variant of logistic regression [55] that uses features extracted from the STFT of the pulse, followed by PCA for dimensionality reduction.

### G. Learning Vector Quantization

The Learning Vector Quantization (LVQ) is a prototype-based supervised classification algorithm [56]. It represents each class by a set of codebook or prototype vectors. During training, these prototypes are moved closer to or further away from the training instances based on whether the classification is correct. A new pulse is classified by assigning it to the class of the nearest prototype.

### H. Support Vector Machine

The Support Vector Machine (SVM) is a supervised learning model used for classification and regression [19]. For classification, it finds an optimal hyperplane that maximizes the margin between the two classes in a high-dimensional feature space. The features used are the tail-to-total charge ratio and the total charge of the pulse.

### I. Tempotron

The Tempotron is a biologically plausible supervised learning algorithm for a single spiking neuron [9]. It is designed to classify patterns based on the precise timing of spikes in the input spike trains. For PSD, the continuous pulse waveform is first encoded into a sequence of spikes. The Tempotron then learns a set of synaptic weights to produce a spike for one class of pulses (e.g., neutrons) and remain sub-threshold for the other class (e.g., gamma-rays).

## VII. COMPARATIVE STUDY

### A. Datasets and Experimental Setups

This study utilized two distinct datasets for PSD. The first dataset was acquired from a $^{241}$Am$^9$Be neutron source. The signals were detected using an EJ299-33 plastic scintillator and digitized by a TPS2000B oscilloscope, yielding a dataset of approximately 10,000 pulses. Corrupted signals, resulting from events such as incomplete energy deposition and signal pile-up, were removed through a preprocessing step. This procedure specifically targeted pulses with artifacts like extended flat peaks and multiple peaks. No filtering or other denoising techniques were applied. After preprocessing, the $^{241}$Am$^9$Be dataset contained 9,414 pulses and represents a typical PSD application where ground-truth labels are unavailable. See more details in [57].

The second dataset was a time-of-flight (TOF) dataset from a $^{238}$Pu$^9$Be source, measured with yttrium aluminum perovskite (YAP:Ce) gamma-ray detectors [58] and an NE213A fast neutron detector [59]. In this dataset, particle labels were verified via coincidence measurements, providing a reliable ground truth for supervised learning [60]. The TOF measurements captured both tagged-neutron and prompt-gamma-ray events. A tagged-neutron event is defined by the detection of a characteristic 4.44 MeV gamma-ray in the YAP:Ce detector in coincidence with a fast neutron in the NE213A detector. A prompt-gamma-ray event refers to time-correlated gamma rays detected in both detectors. Non-prompt gamma events were filtered from the tagged-neutron data using a charge integration threshold. See more details in [61].

The $^{238}$Pu$^9$Be dataset comprises 21,001 pulses after applying a 2 MeV software threshold. For model development and evaluation, this dataset was partitioned into a validation set (80%; 16,802 signals), a training set (18%; 3,778 signals), and a test set (2%; 421 signals). The training set consisted of 1,291 gamma-ray and 2,487 neutron events, while the test set included 144 gamma-ray and 277 neutron events, preserving the natural class distribution of the experimental data. This partitioning strategy, with a large validation set, was designed to establish a robust benchmark for comparing diverse PSD methodologies, as traditional statistical methods require a substantial number of signals for statistically significant analysis.

To accompany this study, we have open-sourced a PSD toolbox, publicly available on GitHub at https://github.com/HaoranLiu507/PulseShapeDiscrimination. The toolbox includes Python implementations for all statistical and prior-knowledge methods, while the MATLAB version contains implementations for the statistical methods. The DL and ML methods were exclusively implemented in Python to leverage its extensive libraries and frameworks. Additionally, we also provide various common pulse signal filters in our PSD toolbox, including Butterworth [62], Chebyshev [63], Elliptic [64], Fourier [65], Least Mean Square (LMS) [66], median [67], morphological [68], moving average [69], wavelet [70], Wiener [71], and windowed-sinc filters [72]. All experiments were conducted on a workstation with an NVIDIA RTX 4090 GPU and an Intel i7-13700K CPU.

All PSD methods in the comparative study adhere to their original architecture and parameter settings as initially proposed. We adjusted only the parameters pertinent to pulse signal characteristics, such as pulse length and the region index of the pulse tail. Moreover, all pulse signals are processed with a moving average filter to reduce noise prior to PSD analysis. Limited by the vast number of methodologies and parameters analyzed in this study, we do not present setting details here. Interested readers are referred to the PSD toolbox we provided, which includes full implementation details.

### B. Comparative Performance of PSD Methodologies

To initiate our comparative analysis, all PSD algorithms were first evaluated on the $^{241}$Am$^9$Be dataset, which consists of 9,414 preprocessed pulses from an EJ299-33 plastic scintillator exposed to a 241Am9Be neutron source. This dataset lacks ground-truth particle labels, representing a typical scenario in PSD applications where supervised classification training is not feasible. Consequently, for statistical methods—including time-domain, frequency-domain, and neural network-based approaches—we directly computed PSD factors for each pulse and assessed performance using the Figure of Merit (FOM), a standard metric that quantifies the separation between neutron and gamma-ray distributions in the PSD factor histogram. The FOM is defined as:

$$FOM = \frac{S}{FWHM_1 + FWHM_2}. \quad (26)$$

where S is the distance between two peak centroids, while $FWHM_1$ and $FWHM_1$ are the Full Width at Half Maximum of two Gaussian distributions. These distributions are fitted from the histogram of PSD factors, as shown in Figure 2. The FOM can be expressed in terms of the means ($\mu_1$, $\mu_2$) and standard deviations ($\sigma_1$, $\sigma_2$) of these two Gaussian distributions as follows:

$$FOM = \frac{|\mu_1 - \mu_2|}{2\sqrt{2\ln(2)}(\sigma_1 + \sigma_2)}. \quad (27)$$

For DL and ML methods, which typically require labeled data for classification tasks, we adapted them to a regression framework. Specifically, models were trained to predict continuous PSD factors that were previously generated by statistical methods. Using a training set of 2,826 signals (30% of the data), the models learned from raw or transformed pulse waveforms (inputs) to predict the statistical PSD factors (outputs). This approach allows us to leverage the feature-learning capabilities of DL and ML while enabling their evaluation on unlabeled data through subsequent FOM computation on the predicted PSD factors. Only methods capable of regression were applied, and for each DL and ML model, we selected the statistical method that yielded the best regression performance during training.

The results of this evaluation are summarized in Table 1, which presents the FOM values for all methods, organized by taxonomy. For DL and ML methods, the table entries indicate the specific statistical method used as the regression target (e.g., CNNDEEP-ZC denotes a CNNDEEP model trained to

TABLE I. FOM VALUES FOR PSD METHODS ON THE 241AM9BE DATASET

| Tax. | Method | | | | | | | |
|---|---|---|---|---|---|---|---|---|
| Time-domain | CC | CI | FEPS | GP | LLR | LMT | PCA | PGA |
| | **1.4761** | 1.2373 | 1.0179 | 1.359 | **1.3787** | 1.2532 | **1.4312** | 0.8015 |
| | PR | ZC | | | | | | |
| | 0.9459 | 1.3041 | | | | | | |
| Frequency-domain | FGA | SDCC | DFT | WT1 | WT2 | FS | SD | |
| | 0.8798 | 0.5697 | **1.6247** | **1.3699** | 1.0239 | Failed | **1.1626** | |
| Neural network | LG | PCNN | RCNN | SCM | | | | |
| | 1.4469 | **1.6104** | **1.8047** | **1.6568** | | | | |
| Deep learning | CNNDEEP-ZC | CNNFT-LMT | CNNSHAL-RCNN | CNNSP-RCNN | CNNSTFT-PCNN | CNNWT-PCNN | ENN-RCNN | GRU-SDCC |
| | 0.7107 | 1.327 | 1.9053 | 2.4126 | 1.6146 | 1.7892 | 1.9412 | **3.8522** |
| | LSTM-FS | MAM-SDCC | MLP1-RCNN | MLP1FT-RCNN | MLP1PCA-RCNN | MLP1STFT-RCNN | MLP1WT-RCNN | MLP2-RCNN |
| | 3.2376 | 1.6221 | 2.4828 | 2.0265 | 2.0453 | 2.5659 | 2.3549 | **3.7273** |
| | MLP2FT-RCNN | MLP2PCA-SCM | MLP2STFT-RCNN | MLP2WT-RCNN | MLP3-RCNN | MLP3FT-SCM | MLP3PCA-RCNN | MLP3STFT-RCNN |
| | 2.3135 | **3.5735** | 2.1115 | 1.9087 | 1.8709 | 3.1381 | 1.8371 | 1.9562 |
| | MLP3WT-RCNN | RNN-RCNN | TRAN-LG | | | | | |
| | 1.8339 | 3.1477 | 1.4863 | | | | | |
| Machine learing | BDT-DFT | DT-RCNN | KNN-RCNN | LINRE-LG | | | | |
| | **2.3536** | **1.7681** | **2.4479** | 1.2245 | | | | |

a. The top three methods in each category are highlighted in **bold**.

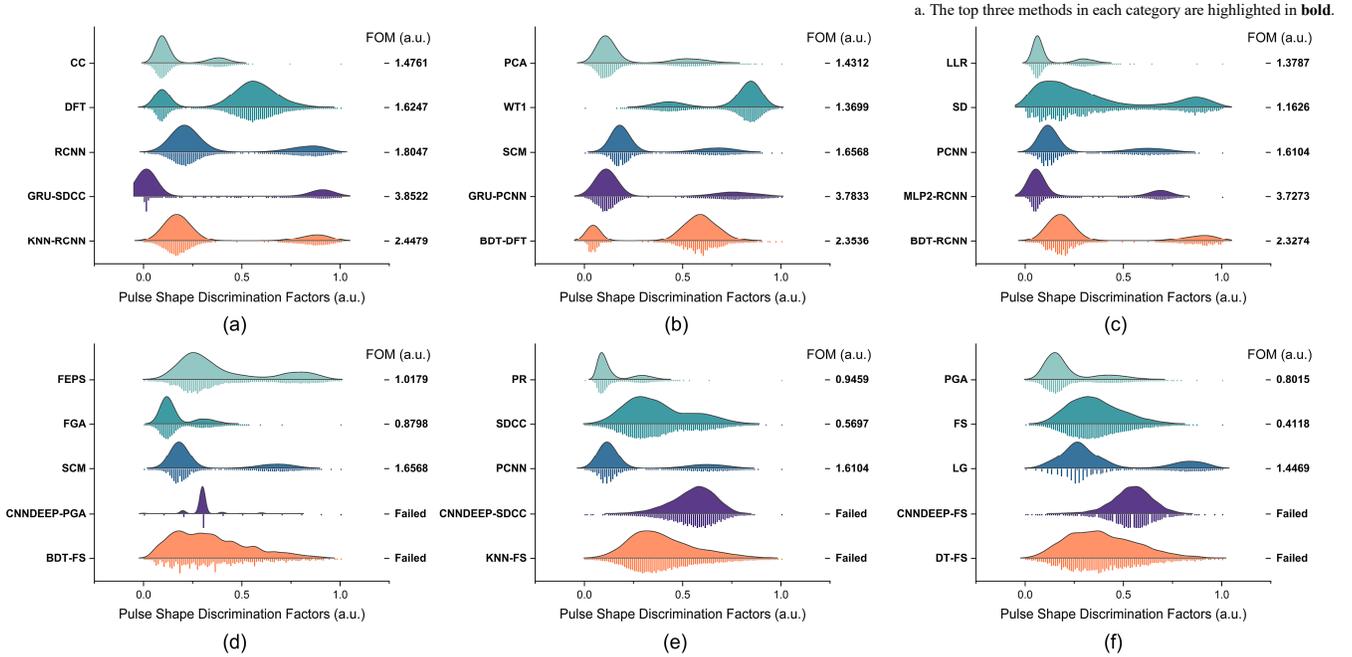

Figure 2. PSD Factor Histograms with Gaussian Fits for Selected High- and Low-Performing Methods on the $^{241}$Am$^9$Be Dataset

regress on PSD factors from the ZC method). Figure 2 complements this by illustrating representative PSD factor histograms for selected methods, with overlaid Gaussian fits and corresponding FOM values, visually demonstrating the separation achieved.

Examining Table 1, we observe a wide range of performances across the methods. In the time-domain category, the CC method achieves the highest FOM of 1.4761, closely followed by PCA at 1.4312, while PGA performs the lowest at 0.8015. Frequency-domain methods show greater variability, with DFT leading at 1.6247 and FS trailing at failure. Neural network methods, which can be viewed as intelligent extensions of statistical approaches, generally perform well, with RCNN topping this subcategory at 1.8047.

The DL methods demonstrate superior performance overall, particularly when regressing on effective statistical targets. The Gated Recurrent Unit (GRU) model, when trained on SDCC factors, achieves the highest FOM of 3.8522 across all methods, highlighting the power of recurrent architectures in capturing temporal dependencies for PSD regression. Other notable performers include the two-hidden-layer MLP (MLP2) at 3.7273 (regressing on RCNN) and the Long Short-Term Memory (LSTM) at 3.2376 (on FS). ML methods also show strong results, with Boosted Decision Tree (BDT) on DFT achieving 2.3536.

Overall, DL methods often double or triple the FOM of statistical baselines, especially when regressing on high-performing targets like RCNN or SCM. This indicates that DL can augment statistical features effectively. Interestingly, strong results from regressing on weaker targets (e.g., GRU on SDCC) suggest DL's ability to uncover hidden patterns. These insights highlight the value of hybrid approaches in PSD. Details on the performance of all PSD methods are available in Supplementary Information Tables S1.

Figure 2 provides visual insight into these performances, showcasing PSD factor histograms for selected methods representing the three best and three worst performers in each category, with overlaid Gaussian fits and corresponding FOM values. For example, high performers like LLR (FOM=1.3787), WT1 (FOM=1.3699), and PCNN (FOM=1.6104) exhibit clear bimodal distributions, while lower performers show overlapping or failed separations (e.g., CNNDEEP-FGA, and DT-FS). The figure illustrates how advanced methods, such as RCNN and DL regressions, produce sharper, more separated peaks compared to simpler statistical approaches. Moreover, it demonstrates the substantial performance gaps between top and bottom methods, particularly evident in frequency-domain, DL, and ML categories, where FOMs range from over 3.0 to below 1.0. In contrast, time-domain and neural network methods display more consistent performance across their respective ranges.

This analysis reveals several key trends. First, while traditional statistical methods provide solid baselines, their performance is often surpassed by DL models that can learn to mimic and enhance these factors through regression. Second, the choice of regression target significantly impacts DL and ML performance; for instance, regressing on lower-performing statistical methods like SDCC or FS can still yield high FOMs when combined with models like GRU or LSTM, suggesting that these models extract additional discriminatory information. Finally, neural network methods (e.g., RCNN, SCM) serve as excellent targets for regression, likely due to their biologically inspired feature extraction aligning well with DL architecture.

Building upon the insights from the $^{241}$Am$^9$Be dataset, we extended our comparative analysis to the $^{238}$Pu$^9$Be dataset. A key advantage of this dataset is the availability of ground-truth particle labels derived from TOF measurements. This enables a direct evaluation of classification performance using a suite of standard metrics, complementing the FOM analysis used for the $^{241}$Am$^9$Be data.

To comprehensively assess classification performance, several key metrics are employed. Accuracy provides a straightforward measure of the overall correctness of a model's predictions. However, in contexts such as PSD where class imbalances may exist, accuracy alone can be insufficient. Therefore, we also utilize precision and recall for a more granular analysis. Precision measures the proportion of true positive predictions among all positive predictions, indicating the exactness of the classifier. Recall, or sensitivity, measures the proportion of actual positives that were correctly identified, indicating the completeness of the classifier. The F1-score is then used to provide a single, summary metric. As the harmonic mean of precision and recall, it offers a balanced assessment of a model's performance, which is particularly useful for comparing models in a generalized manner.

To ensure a comprehensive assessment, all methods were evaluated in paradigms appropriate to their design. For methods that produce a continuous PSD factor, either directly via statistical analysis or through a regression model, classification was performed using a non-parametric KNN classifier. This approach assigns a label to each pulse based on the majority vote of its five nearest neighbors in the PSD factor space, avoiding the potential biases of manual thresholding. In parallel, DL and machine learning ML models were also trained for direct binary classification. This dual evaluation strategy allows for quantifying both the classification performance of all methods on a level playing field and the distribution separation (via FOM) for regression-based techniques.

The detailed results are presented in Table 2, organized by taxonomy. The left columns detail statistical approaches classified via KNN and direct DL/ML classifiers. Performance is reported with classification metrics and the FOM, where an FOM value below 0.5 is marked as "Failed" and is Not Applicable (N/A) for classifiers that output probabilities. The right columns showcase hybrid regression models. For each DL and ML model, the table presents the results from the optimal pairing with a statistical feature extractor as the regression target (e.g., CNNDEEP-SCM indicates SCM was the best-performing target for the CNNDEEP model). For these hybrids, both classification metrics from the predicted factors and the resulting FOM are reported.

Table 2 highlights the efficacy of various methods, evaluated through several metrics. Here we focus on F1-score, a balanced measure of classification performance, and FOM, which quantifies histogram separation. Statistical methods (TD, FD, NN) show a broad performance range, with F1-scores spanning 0.590 (FS) to 0.953 (CI) and FOMs from Failed (WT1) to 0.9415 (LG). Prior-knowledge methods (DL, ML) generally outperform, achieving F1-scores up to 0.962 (MLP2STFT) as direct classifiers and 0.958 (MLP1WT-CI) as hybrid regression models, with GRU-CI yielding an exceptional FOM of 1.8531.

In the TD category, CC and CI excel with F1-scores of 0.951 and 0.953 and FOMs of 0.9240 and 0.9193,

TABLE II. CLASSIFICATION PERFORMANCE METRICS AND FOM FOR PSD METHODS ON THE $^{238}$PU$^9$BE DATASET

| Tax. | Method | Criteria | | | | | Method | Criteria | | | | |
|---|---|---|---|---|---|---|---|---|---|---|---|---|
| | | *Acc.* | *Prec.* | *Rec.* | *F1* | *FOM* | | *Acc.* | *Prec.* | *Rec.* | *F1* | *FOM* |
| TD | **CC** | **0.951** | **0.951** | **0.951** | **0.951** | **0.9240** | **CI** | **0.953** | **0.953** | **0.953** | **0.953** | **0.9193** |
| | FEPS | 0.838 | 0.836 | 0.838 | 0.837 | 0.5582 | GP | 0.644 | 0.633 | 0.644 | 0.637 | 0.5580 |
| | LLR | 0.779 | 0.779 | 0.779 | 0.779 | Failed | LMT | 0.797 | 0.794 | 0.797 | 0.790 | Failed |
| | PCA | 0.831 | 0.828 | 0.831 | 0.826 | Failed | **PGA** | **0.864** | **0.863** | **0.864** | **0.863** | **Failed** |
| | PR | 0.633 | 0.615 | 0.633 | 0.621 | Failed | ZC | 0.858 | 0.856 | 0.858 | 0.856 | Failed |
| FD | **DFT** | **0.932** | **0.932** | **0.932** | **0.931** | **0.5461** | **FGA** | **0.932** | **0.932** | **0.932** | **0.932** | **0.7565** |
| | FS | 0.609 | 0.582 | 0.609 | 0.590 | Failed | SD | 0.810 | 0.807 | 0.810 | 0.808 | Failed |
| | **SDCC** | **0.894** | **0.894** | **0.894** | **0.893** | **0.6294** | WT1 | 0.717 | 0.721 | 0.717 | 0.719 | Failed |
| | WT2 | 0.675 | 0.663 | 0.675 | 0.667 | 0.6282 | | | | | | |
| NN | LG | 0.685 | 0.669 | 0.685 | 0.670 | 0.9415 | **PCNN** | **0.948** | **0.948** | **0.948** | **0.948** | **0.9155** |
| | **RCNN** | **0.952** | **0.952** | **0.952** | **0.952** | **0.7195** | SCM | 0.949 | 0.949 | 0.949 | 0.949 | 0.6282 |
| DL | CNNDEEP | 0.658 | 0.433 | 0.658 | 0.522 | N/A | CNNDEEP-SCM | 0.924 | 0.924 | 0.924 | 0.924 | 0.8049 |
| | CNNFT | 0.943 | 0.944 | 0.943 | 0.943 | N/A | CNNFT-RCNN | 0.949 | 0.949 | 0.949 | 0.949 | 0.9105 |
| | CNNSHAL | 0.955 | 0.955 | 0.955 | 0.955 | N/A | CNNSHAL-PCNN | 0.940 | 0.941 | 0.941 | 0.941 | 0.9002 |
| | CNNSP | 0.953 | 0.953 | 0.953 | 0.953 | N/A | CNNSP-PCNN | 0.950 | 0.950 | 0.951 | 0.951 | 1.0751 |
| | CNNSTFT | 0.342 | 0.117 | 0.342 | 0.174 | N/A | CNNSTFT-CI | 0.949 | 0.949 | 0.949 | 0.949 | 0.9306 |
| | CNNWT | 0.658 | 0.433 | 0.658 | 0.522 | N/A | CNNWT-CI | 0.953 | 0.953 | 0.954 | 0.954 | 0.9857 |
| | ENN | 0.945 | 0.945 | 0.945 | 0.945 | N/A | ENN-RCNN | 0.953 | 0.953 | 0.953 | 0.953 | 1.0167 |
| | GRU | 0.940 | 0.941 | 0.940 | 0.940 | N/A | GRU-CI | 0.937 | 0.937 | 0.937 | 0.937 | 1.8531 |
| | LSTM | 0.775 | 0.780 | 0.775 | 0.755 | N/A | LSTM-LMT | 0.683 | 0.681 | 0.696 | 0.696 | 1.0933 |
| | MAM | 0.658 | 0.433 | 0.658 | 0.522 | N/A | MAM-FGA | 0.933 | 0.934 | 0.933 | 0.933 | 0.8364 |
| | MLP1 | 0.954 | 0.955 | 0.954 | 0.953 | N/A | MLP1-SCM | 0.956 | 0.956 | 0.956 | 0.956 | 1.0238 |
| | MLP1FT | 0.952 | 0.952 | 0.952 | 0.952 | N/A | MLP1FT-CC | 0.950 | 0.950 | 0.951 | 0.951 | 0.9943 |
| | MLP1PCA | 0.956 | 0.956 | 0.956 | 0.956 | N/A | MLP1PCA-CI | 0.954 | 0.954 | 0.954 | 0.954 | 0.9998 |
| | **MLP1STFT** | **0.962** | **0.962** | **0.962** | **0.962** | **N/A** | MLP1STFT-CI | 0.952 | 0.952 | 0.952 | 0.952 | 1.0346 |
| | MLP1WT | 0.948 | 0.950 | 0.948 | 0.947 | N/A | MLP1WT-CI | 0.957 | 0.958 | 0.958 | 0.958 | 1.0021 |
| | MLP2 | 0.929 | 0.929 | 0.929 | 0.928 | N/A | MLP2-SCM | 0.919 | 0.918 | 0.919 | 0.919 | 1.0307 |
| | MLP2FT | 0.947 | 0.947 | 0.947 | 0.947 | N/A | MLP2FT-CI | 0.951 | 0.951 | 0.951 | 0.951 | 0.9495 |
| | MLP2PCA | 0.956 | 0.956 | 0.956 | 0.956 | N/A | MLP2PCA-CI | 0.954 | 0.954 | 0.954 | 0.954 | 0.9461 |
| | **MLP2STFT** | **0.963** | **0.963** | **0.963** | **0.962** | **N/A** | MLP2STFT-RCNN | 0.951 | 0.951 | 0.952 | 0.952 | 1.0081 |
| | MLP2WT | 0.930 | 0.930 | 0.930 | 0.930 | N/A | MLP2WT-CC | 0.951 | 0.951 | 0.951 | 0.951 | 0.9396 |
| | MLP3 | 0.956 | 0.956 | 0.956 | 0.956 | N/A | MLP3-RCNN | 0.955 | 0.955 | 0.955 | 0.955 | 0.9339 |
| | MLP3FT | 0.954 | 0.955 | 0.954 | 0.954 | N/A | MLP3FT-CI | 0.953 | 0.953 | 0.953 | 0.953 | 0.9855 |
| | MLP3PCA | 0.937 | 0.937 | 0.937 | 0.937 | N/A | MLP3PCA-CI | 0.953 | 0.953 | 0.954 | 0.954 | 0.9352 |
| | **MLP3STFT** | **0.961** | **0.961** | **0.961** | **0.961** | **N/A** | MLP3STFT-CI | 0.954 | 0.954 | 0.954 | 0.954 | 0.9392 |

| Tax. | Method | Criteria | | | | | Method | Criteria | | | | |
|---|---|---|---|---|---|---|---|---|---|---|---|---|
| | | Acc. | Prec. | Rec. | F1 | FOM | | Acc. | Prec. | Rec. | F1 | FOM |
| | MLP3WT | 0.933 | 0.934 | 0.933 | 0.933 | N/A | MLP3WT-CI | 0.954 | 0.954 | 0.954 | 0.954 | 0.9361 |
| | RNN | 0.658 | 0.433 | 0.658 | 0.522 | N/A | RNN-RCNN | 0.744 | 0.747 | 0.755 | 0.755 | 1.2578 |
| | TRAN | 0.944 | 0.944 | 0.944 | 0.944 | N/A | TRAN-FS | 0.930 | 0.931 | 0.931 | 0.931 | 0.7113 |
| ML | BDT | 0.925 | 0.925 | 0.925 | 0.925 | N/A | **BDT-PCNN** | **0.947** | **0.947** | **0.947** | **0.947** | **1.1468** |
| | DT | 0.652 | 0.652 | 0.652 | 0.652 | N/A | DT-CC | 0.694 | 0.682 | 0.694 | 0.684 | 0.9255 |
| | FCM | 0.689 | 0.795 | 0.689 | 0.694 | N/A | GMM | 0.940 | 0.940 | 0.940 | 0.940 | N/A |
| | **KNN** | **0.948** | **0.948** | **0.948** | **0.948** | N/A | **KNN-PCNN** | **0.944** | **0.944** | **0.944** | **0.944** | **1.1016** |
| | LINRE-FEPS | 0.630 | 0.606 | 0.630 | 0.612 | 0.6300 | LOGRE | 0.658 | 0.433 | 0.658 | 0.522 | N/A |
| | LRSTFT | 0.942 | 0.941 | 0.942 | 0.941 | N/A | LVQ | 0.877 | 0.886 | 0.877 | 0.879 | N/A |
| | SVM | 0.936 | 0.935 | 0.936 | 0.935 | N/A | TEM | 0.812 | 0.813 | 0.812 | 0.813 | N/A |

a. The top three methods in each category are highlighted in **bold**.

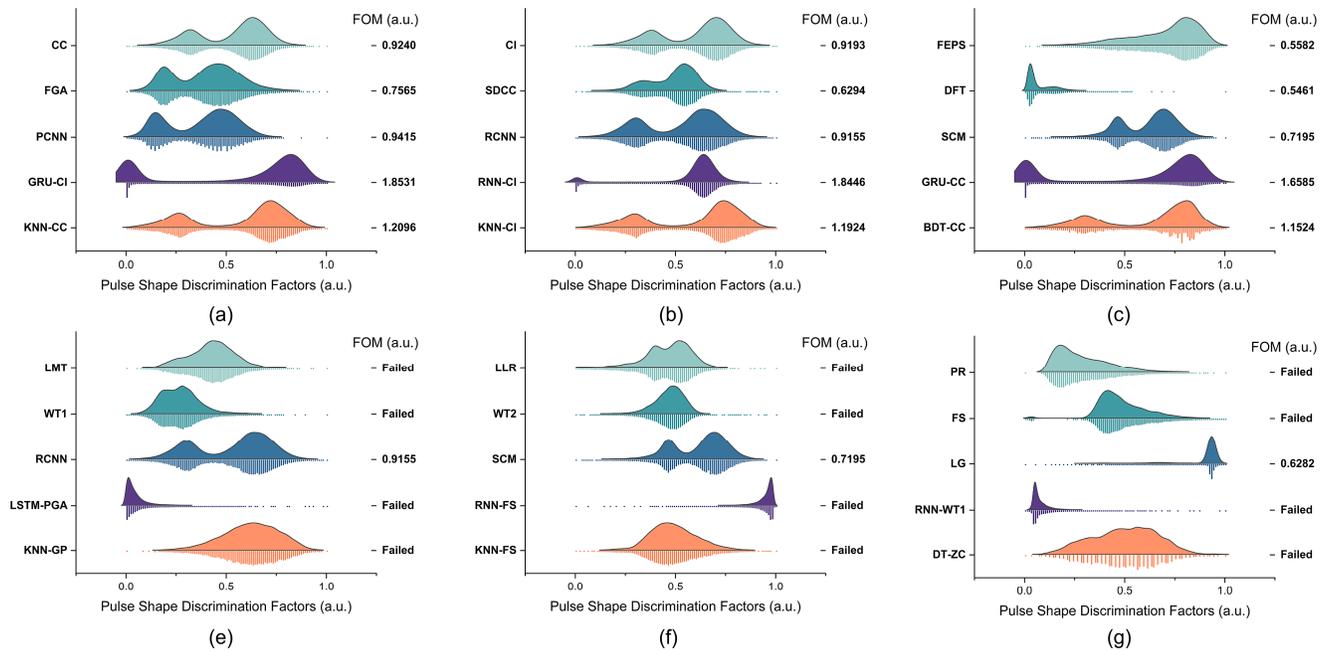

Figure 3. PSD Factor Histograms with Gaussian Fits for Selected High- and Low-Performing Methods on the $^{238}$Pu$^9$Be Dataset

respectively, reflecting consistent performance. In contrast, GP and PR lag with F1-scores of 0.637 and 0.621 and FOMs of 0.5580 and Failed, indicating weaker discriminative feature capture.

FD methods also exhibit variability: DFT and FGA achieve F1-scores of 0.931–0.932 but differ in FOMs (0.5461 vs. 0.7565), while WT1 records an F1-score of 0.719 and a Failed FOM. In contrast, NN methods perform robustly, with RCNN (F1=0.952, FOM=0.7195) and PCNN (F1=0.948, FOM=0.9155) demonstrating strong and consistent results.

DL and ML classification models frequently achieve high F1-scores, with MLP2STFT (0.962), MLP1STFT (0.962), and KNN (0.948) performing exceptionally well. However, some models, such as CNNSTFT, underperform with an F1-score of 0.174. In contrast, regression models also demonstrated strong performance in PSD tasks. GRU-CI led with an FOM of 1.8531 and an F1-score of 0.937, while MLP1WT-CI achieved an F1-score of 0.958 and an FOM of 1.0021. Comprehensive regression model performance details are available in Supplementary Information Tables S3 and S4.

Figure 3 visually corroborates these findings, with top performers like KNN-CC showing precise probability distributions and GRU-CI displaying sharply separated factors (high FOM), while poor performance like WT2 and PR exhibit only a single distribution.

A critical observation is that F1-score more accurately reflects true model performance than FOM, as it directly accounts for classification accuracy against ground-truth labels, whereas FOM relies on histogram separation and assumes Gaussian distributions. Differences arise when models produce well-separated but misaligned distributions (high FOM, low F1) or accurate classifications with overlapping histograms (high F1, low FOM). For instance, RNN-RCNN hybrid achieves F1=0.755 but an exceptional FOM=1.2578, suggesting strong separation yet classification errors due to threshold sensitivity. Conversely, TRAN-FS attains high F1=0.931 but low FOM=0.7113, indicating superior classification despite poor histogram separation.

TABLE III.  AVERAGE CLASSIFICATION PERFORMANCE AND COMPUTATIONAL COSTS OF REGRESSORS ACROSS ALL FEATURE EXTRACTORS

| Tax. | Method | Criteria | | | | | | Method | Criteria | | | | | |
|---|---|---|---|---|---|---|---|---|---|---|---|---|---|---|
| | | *Acc.* | *Prec.* | *Rec.* | *F1* | *Para.* | *FLOPs* | | *Acc.* | *Prec.* | *Rec.* | *F1* | *Para.* | *FLOPs* |
| DL | CNNDEEP | 0.796 ±0.101 | 0.785 ±0.117 | 0.796 ±0.101 | 0.785 ±0.118 | $2.69 \times 10^7$ | $2.61 \times 10^9$ | CNNFT | 0.809 ±0.123 | 0.803 ±0.131 | 0.809 ±0.123 | 0.797 ±0.139 | $6.45 \times 10^6$ | $2.23 \times 10^8$ |
| | CNNSHAL | 0.924 ±0.011 | 0.924 ±0.011 | 0.924 ±0.011 | 0.923 ±0.012 | 7505 | $2.67 \times 10^6$ | CNNSP | 0.851 ±0.110 | 0.844 ±0.125 | 0.851 ±0.110 | 0.843 ±0.124 | $6.45 \times 10^6$ | $2.23 \times 10^8$ |
| | CNNSTFT | 0.851 ±0.107 | 0.847 ±0.113 | 0.851 ±0.107 | 0.847 ±0.112 | $6.45 \times 10^6$ | $2.23 \times 10^8$ | CNNWT | 0.840 ±0.117 | 0.836 ±0.123 | 0.840 ±0.117 | 0.837 ±0.121 | $6.45 \times 10^6$ | $2.23 \times 10^8$ |
| | ENN | 0.839 ±0.106 | 0.836 ±0.111 | 0.839 ±0.106 | 0.836 ±0.110 | $8.10 \times 10^4$ | $1.48 \times 10^5$ | GRU | 0.813 ±0.140 | 0.803 ±0.154 | 0.813 ±0.140 | 0.805 ±0.150 | 226 | $1.05 \times 10^5$ |
| | LSTM | 0.605 ±0.025 | 0.574 ±0.029 | 0.605 ±0.025 | 0.583 ±0.027 | 266 | $1.60 \times 10^5$ | MAM | 0.903 ±0.068 | 0.901 ±0.074 | 0.903 ±0.068 | 0.901 ±0.072 | 5297 | $1.35 \times 10^6$ |
| | MLP1 | 0.846 ±0.108 | 0.844 ±0.112 | 0.846 ±0.108 | 0.844 ±0.111 | 1002 | 1001 | MLP1FT | 0.880 ±0.098 | 0.878 ±0.103 | 0.880 ±0.098 | 0.878 ±0.101 | 503 | $5.04 \times 10^4$ |
| | MLP1PCA | 0.835 ±0.111 | 0.833 ±0.115 | 0.835 ±0.111 | 0.833 ±0.114 | 101 | $1.00 \times 10^5$ | MLP1STFT | 0.831 ±0.126 | 0.827 ±0.131 | 0.831 ±0.126 | 0.827 ±0.130 | 1057 | $6.06 \times 10^4$ |
| | MLP1WT | 0.846 ±0.106 | 0.844 ±0.110 | 0.846 ±0.106 | 0.844 ±0.108 | 1003 | 3002 | MLP2 | 0.884 ±0.062 | 0.882 ±0.067 | 0.884 ±0.062 | 0.882 ±0.066 | 191 | 170 |
| | MLP2FT | 0.858 ±0.100 | 0.856 ±0.103 | 0.858 ±0.100 | 0.856 ±0.102 | 5151 | $5.50 \times 10^4$ | MLP2PCA | 0.826 ±0.115 | 0.822 ±0.121 | 0.826 ±0.115 | 0.822 ±0.119 | 1131 | $1.01 \times 10^5$ |
| | MLP2STFT | 0.836 ±0.111 | 0.839 ±0.109 | 0.836 ±0.111 | 0.827 ±0.127 | $1.07 \times 10^4$ | $7.02 \times 10^4$ | MLP2WT | 0.826 ±0.107 | 0.823 ±0.111 | 0.826 ±0.107 | 0.823 ±0.110 | $1.02 \times 10^4$ | $1.21 \times 10^4$ |
| | MLP3 | 0.825 ±0.114 | 0.821 ±0.120 | 0.825 ±0.114 | 0.822 ±0.118 | $6.88 \times 10^5$ | $6.87 \times 10^5$ | MLP3FT | 0.874 ±0.084 | 0.872 ±0.087 | 0.874 ±0.084 | 0.872 ±0.087 | $4.33 \times 10^5$ | $4.82 \times 10^5$ |
| | MLP3PCA | 0.822 ±0.114 | 0.819 ±0.119 | 0.822 ±0.114 | 0.819 ±0.117 | $2.27 \times 10^5$ | $3.26 \times 10^5$ | MLP3STFT | 0.830 ±0.111 | 0.827 ±0.114 | 0.830 ±0.111 | 0.827 ±0.114 | $7.16 \times 10^5$ | $7.75 \times 10^5$ |
| | MLP3WT | 0.824 ±0.113 | 0.821 ±0.118 | 0.824 ±0.113 | 0.821 ±0.116 | $6.89 \times 10^5$ | $6.90 \times 10^5$ | RNN | 0.643 ±0.064 | 0.616 ±0.076 | 0.643 ±0.064 | 0.622 ±0.069 | 146 | $6.02 \times 10^4$ |
| | TRAN | 0.855 ±0.116 | 0.850 ±0.127 | 0.855 ±0.116 | 0.851 ±0.124 | $5.63 \times 10^5$ | $3.36 \times 10^7$ | | | | | | | |
| | BDT | 0.884 ±0.060 | 0.883 ±0.062 | 0.884 ±0.060 | 0.883 ±0.061 | 7244 | 598 | DT | 0.667 ±0.027 | 0.652 ±0.028 | 0.667 ±0.027 | 0.655 ±0.028 | 6923 | 4034 |
| | KNN | 0.861 ±0.098 | 0.859 ±0.101 | 0.861 ±0.098 | 0.859 ±0.100 | N/A | 299 | LINRE | 0.615 ±0.011 | 0.586 ±0.012 | 0.615 ±0.011 | 0.594 ±0.011 | 2005 | 4008 |

This analysis underscores that traditional FOM is not always reliable, particularly when the dataset becomes challenging, as it may overemphasize separation without ensuring correct classifications. Standard metrics like F1-score provide a more robust evaluation in labeled scenarios, advocating for their prioritization in PSD assessments.

*C. Performance of Regressors and Feature Extractors*

The preceding analysis demonstrated that training DL and ML models as regressors on PSD factors from statistical methods is a viable strategy. It was observed that regressor performance is significantly influenced by the chosen feature extractor. This section further investigates this interplay through a two-part analysis. First, we evaluate the average performance of each regressor across all feature extractors to identify models that exhibit both high accuracy and stability, with results presented in Table 3. Second, we assess the average performance of each feature extractor across all regressors to determine which statistical methods are most suitable as learning targets for DL models, shown in Table 4.

TABLE IV. AVERAGE CLASSIFICATION PERFORMANCE OF STATISTICAL FEATURE EXTRACTORS ACROSS ALL REGRESSORS

| Feature-Extractor | Criteria | | | | Feature-Extractor | Criteria | | | |
|---|---|---|---|---|---|---|---|---|---|
| | Acc. | Prec. | Rec. | F1 | | Acc. | Prec. | Rec. | F1 |
| CC | 0.906 ±0.094 | 0.904 ±0.098 | 0.906 ±0.094 | 0.899 ±0.112 | CI | 0.919 ±0.084 | 0.917 ±0.091 | 0.919 ±0.084 | 0.917 ±0.089 |
| DFT | 0.895 ±0.087 | 0.892 ±0.096 | 0.895 ±0.087 | 0.892 ±0.093 | FEPS | 0.857 ±0.109 | 0.853 ±0.120 | 0.857 ±0.109 | 0.854 ±0.117 |
| FGA | 0.907 ±0.066 | 0.906 ±0.071 | 0.907 ±0.066 | 0.906 ±0.070 | FS | 0.712 ±0.115 | 0.700 ±0.124 | 0.712 ±0.115 | 0.703 ±0.121 |
| GP | 0.697 ±0.108 | 0.684 ±0.116 | 0.697 ±0.108 | 0.688 ±0.114 | LG | 0.698 ±0.110 | 0.683 ±0.119 | 0.698 ±0.110 | 0.687 ±0.117 |
| LLR | 0.780 ±0.096 | 0.770 ±0.112 | 0.780 ±0.096 | 0.769 ±0.114 | LMT | 0.805 ±0.051 | 0.801 ±0.054 | 0.805 ±0.051 | 0.798 ±0.054 |
| PCA | 0.791 ±0.095 | 0.783 ±0.106 | 0.791 ±0.095 | 0.783 ±0.103 | PCNN | 0.918 ±0.078 | 0.916 ±0.084 | 0.918 ±0.078 | 0.916 ±0.083 |
| PGA | 0.855 ±0.109 | 0.850 ±0.120 | 0.855 ±0.109 | 0.852 ±0.117 | PR | 0.736 ±0.107 | 0.724 ±0.116 | 0.736 ±0.107 | 0.727 ±0.113 |
| RCNN | 0.912 ±0.088 | 0.910 ±0.094 | 0.912 ±0.088 | 0.910 ±0.093 | SCM | 0.916 ±0.092 | 0.913 ±0.101 | 0.916 ±0.092 | 0.914 ±0.098 |
| SD | 0.840 ±0.092 | 0.836 ±0.100 | 0.840 ±0.092 | 0.837 ±0.097 | SDCC | 0.877 ±0.080 | 0.874 ±0.087 | 0.877 ±0.080 | 0.874 ±0.086 |
| WT1 | 0.777 ±0.096 | 0.775 ±0.104 | 0.777 ±0.096 | 0.774 ±0.101 | WT2 | 0.746 ±0.096 | 0.743 ±0.103 | 0.746 ±0.096 | 0.735 ±0.109 |

Table 3 summarizes the mean classification performance of the DL and ML regressors, averaged across all statistical feature extractors. The evaluation employs standard classification metrics along with computational costs, including parameter counts and Floating-Point Operations (FLOPs), to assess model complexity. By reporting performance as a mean with standard deviation, we can gauge both the overall efficacy and the stability of each regressor.

The analysis of Table 3 reveals a clear performance hierarchy. Within the DL category, CNNSHAL emerges as the top performer (with two 1D CNN and two MLP layers), achieving the highest average F1-score (0.923 ±0.012) and accuracy (0.924 ±0.011). Its exceptionally low standard deviation signifies remarkable stability, indicating that its performance is robust regardless of the feature extractor used. Other strong DL models include MAM and MLP2, which also deliver high F1-scores (0.901 ±0.072 and 0.882 ±0.066, respectively) with low variance, suggesting consistent generalization. In contrast, models like LSTM exhibit weaker and less stable performance (F1-score of 0.583 ±0.027), likely due to difficulties in capturing salient temporal features when paired with less informative extractors. Among the ML models, BDT demonstrates the best results, with an F1-score of 0.883 ±0.061 and moderate stability, substantially outperforming simpler models like DT (0.655 ±0.028) and LINRE (0.594 ±0.011). From a computational standpoint, a clear trade-off exists: simpler architectures like MLP1 and RNN are highly efficient but less powerful, whereas complex models like MAM achieve superior performance at a significant computational cost.

Shifting the focus to the feature extractors, Table 4 presents their average performance when used as regression targets for all the DL and ML models. This aggregated view helps identify which statistical methods function as the most effective "teachers" by consistently producing discriminative PSD factors that are conducive to learning.

The results in Table 4 indicate that the CI method is the premier feature extractor, achieving the highest average F1-score (0.917 ±0.089) and accuracy (0.919 ±0.084). Its success suggests that it generates a robust and well-separated feature space that is highly amenable to regression. The NN extractors PCNN and SCM also perform exceptionally well, with F1-scores of 0.916 ±0.083 and 0.914 ±0.098, respectively. Their strong, stable performance is likely attributable to their biologically inspired feature extraction capabilities, which align well with the learning paradigms of DL models. Methods such as CC (F1-score of 0.899 ±0.112) and RCNN (F1-score of 0.910 ±0.093) are also reliable targets. Conversely, methods like GP (0.688 ±0.114) and FS (0.703 ±0.121) yield lower average scores and greater variability, indicating they produce less discriminative factors that pose a greater challenge for regressors. This analysis confirms that certain TD (CI) and NN (PCNN, SCM, RCNN) statistical methods are the most suitable for generating regression targets in PSD contexts. These findings strongly advocate for hybrid approaches that pair high-performing statistical extractors with advanced DL regressors to maximize overall discrimination efficacy.

### D. Additional Metrics for PSD Performance Evaluation

In the previous section, we demonstrated that the F1-score is more reliable than FOM as an evaluation metric for PSD. However, the F1-score relies on predicted labels derived from a KNN classifier for fairness. To assess classifier performance more generally, without dependence on KNN, we turn to the Receiver Operating Characteristic (ROC) curve—a standardized metric in classification tasks.

The ROC curve plots the True Positive Rate (TPR) against the False Positive Rate (FPR) at various threshold settings, providing a comprehensive view of a model's ability to discriminate between classes across all possible decision boundaries. The Area Under the Curve (AUC) quantifies overall performance, with values closer to 1 indicating superior discrimination.

Based on the AUC values, Figure 4 presents the ROC curves for the best and worst performing DL classifiers in each category: CNN (best: CNNSP, worst: CNNSTFT), RNN

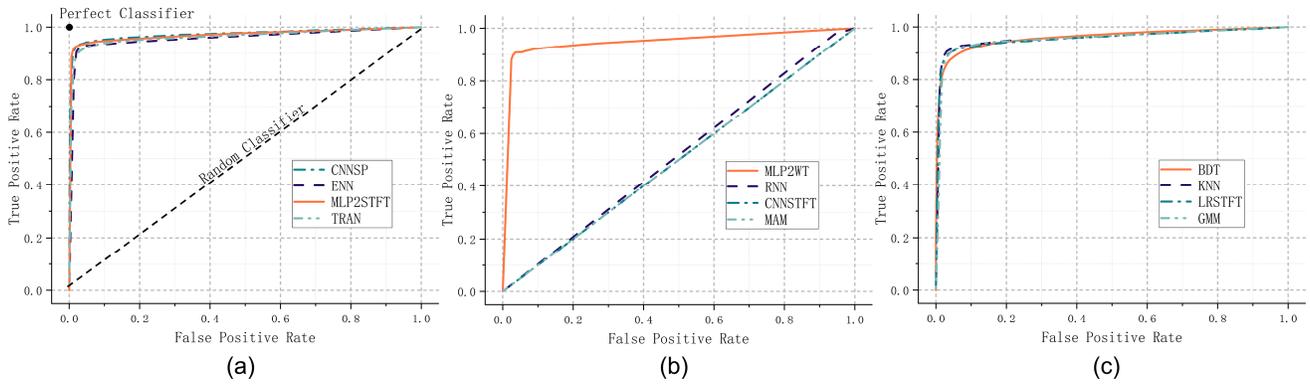

Figure 4. ROC Curves for (a) Best- and (b) Worst-Performing DL and (c) Best-performing ML Classifiers on the $^{238}Pu^9Be$ Dataset

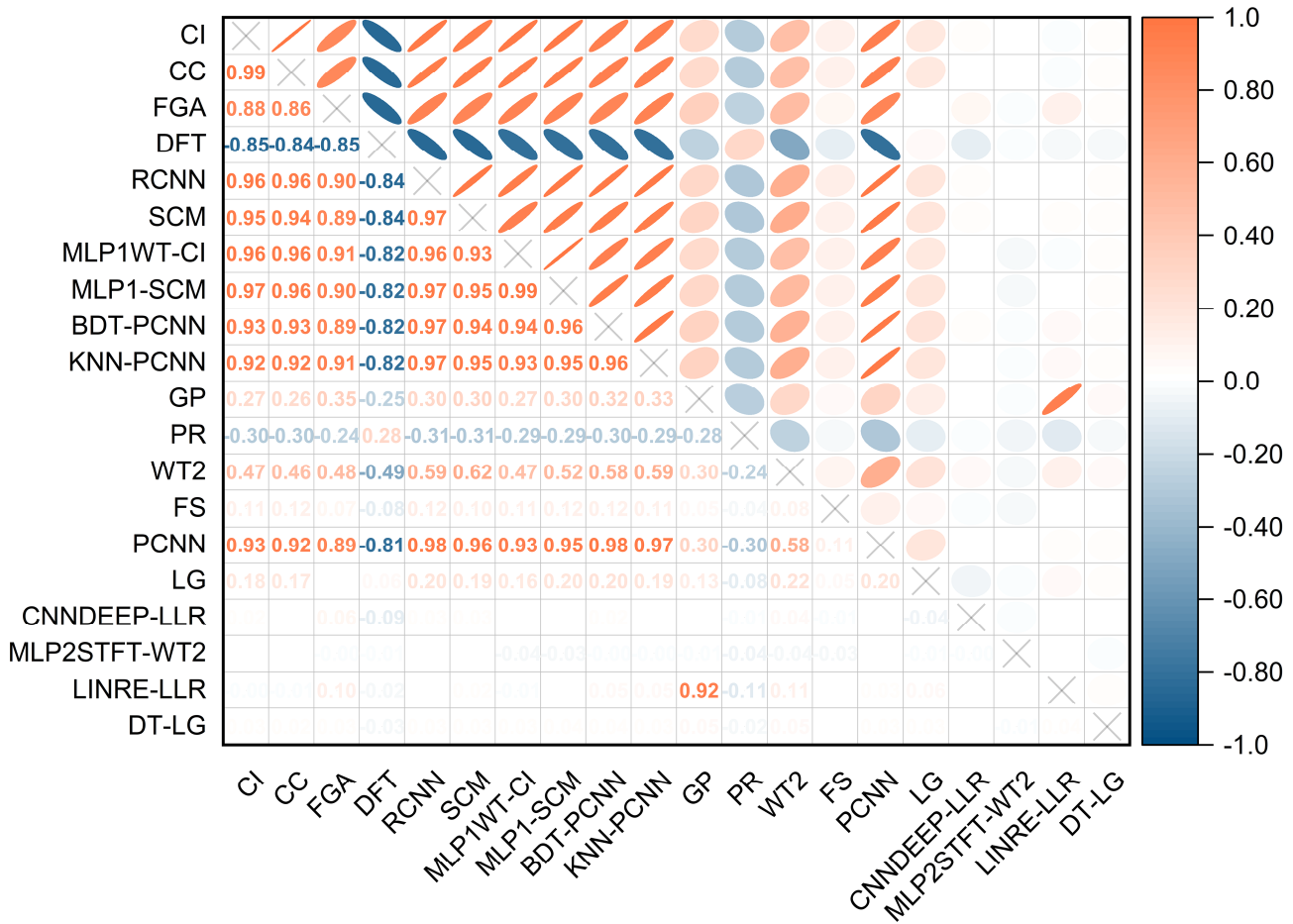

Figure 5. Correlation Heatmap of PSD Factors from High- and Low-Performing Methods on the $^{238}Pu^9Be$ Dataset

(best: ENN, worst: RNN), MLP (best: MLP2STFT, worst: MLP2WT), and advanced networks (best: TRAN, worst: MAM). Additionally, it includes curves for the four best-performing ML classifiers: BDT, GMM, KNN, and LRSTFT.

This figure illustrates the trade-offs between sensitivity and specificity, highlighting how top performers like MLP2STFT achieve near-perfect AUC, while underperformers exhibit curves closer to the diagonal, indicating limited discriminatory power.

The F1-score, ROC curve, and AUC serve as robust metrics for PSD evaluation when ground-truth labels are available. However, practical PSD applications often require evaluation methods independent of ground-truth labels. Consequently, we propose using the correlation between PSD factors from various methods as a supplementary evaluation criterion alongside FOM.

Here, we present a correlation heatmap (Figure 5) demonstrating the correlations between PSD factors from selected high-performing and low-performing methods. We calculated the Pearson correlation coefficients between their PSD factors on the $^{238}$Pu$^9$Be dataset.

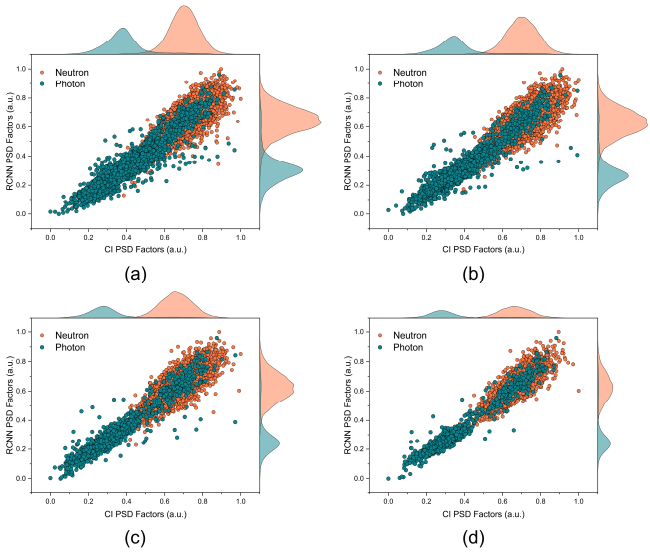

Figure 6. Division of the $^{238}$Pu$^9$Be Dataset into Software Energy Thresholds: (a) 2 MeV, (b) 3 MeV, (c) 4 MeV, and (d) 5 MeV

Based on Figure 5, it is evident that high-performance methods exhibit consistent and strong correlations with each other, either positive or negative, indicating shared discriminatory features. Notably, both positive and negative correlations with effective PSD methods can signify good performance, as different methods may represent pulse signal features inversely; for instance, some assign larger PSD factors to pulses with longer decay times, while others do the inverse. In contrast, correlations drastically decrease when comparing these to poor-performing methods. This suggests that effective PSD methods operate in similar feature spaces, capturing analogous pulse characteristics that enable reliable discrimination, whereas underperforming methods diverge, potentially extracting less relevant or noisy features.

*E. Impact of Energy Thresholds on PSD Performance*

Practical PSD applications often need to process pulse signals under different energy conditions. A broad energy range typically leads to more challenging PSD tasks. Consequently, the performance of PSD algorithms across different energy ranges is crucial for the PSD community. We divided the PuBe dataset into four software energy thresholds: 2 MeV (21,001 signals), 3 MeV (16,520 signals), 4 MeV (11,434 signals), and 5 MeV (5,537 signals), as shown in Figure 6. The same partitioning strategy, 80% validation, 18% training, and 2% testing, was universally applied to each subset. All PSD methodologies were tested on the validation sets, with performance presented in Table 5 and Figure 7.

Table 5 presents the statistical methods and the top two

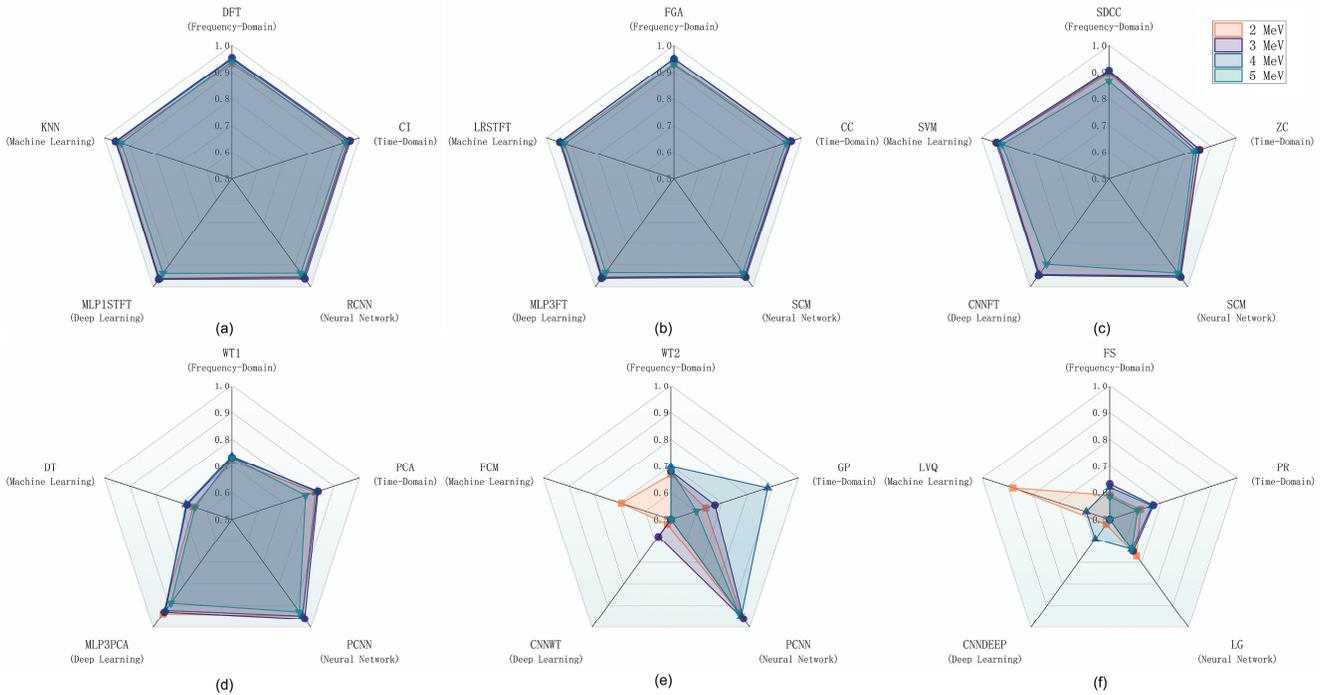

Figure 7. Radar Chart of PSD Method Performance Across Different Energy Thresholds

performing methods for each prior-knowledge-based modality. In general, as the software energy threshold increases from 2 MeV to 4 MeV, the performance of most PSD methods improves, with metrics often rising by 0.01 to 0.03 points, reflecting better discrimination capabilities at higher energies where signal characteristics are more distinct. For example, in the TD, CI's F1-score improves from 0.9529 at 2 MeV to 0.9636 at 3 MeV and 0.9574 at 4 MeV, before a slight drop to 0.9434 at 5 MeV. However, a consistent performance drop is observed at the 5 MeV threshold for all methods, typically by 0.01 to 0.02 points compared to 4 MeV. This decline is attributed to non-prompt gamma events mislabeled during the TOF experiment, an issue that becomes more pronounced as the energy threshold rises and the total

TABLE V. PERFORMANCE METRICS OF SELECTED PSD METHODS ACROSS DIFFERENT ENERGY THRESHOLDS ON THE $^{238}$Pu$^9$Be DATASET

| Tax. | Method | Criteria | Energy | | | | Method | Criteria | Energy | | | |
| --- | --- | --- | --- | --- | --- | --- | --- | --- | --- | --- | --- | --- |
| | | | 2 MeV | 3 MeV | 4 MeV | 5 MeV | | | 2 MeV | 3 MeV | 4 MeV | 5 MeV |
| TD | CI | Acc. | 0.9532 | 0.9639 | 0.9581 | 0.9444 | CC | Acc. | 0.9510 | 0.9607 | 0.9563 | 0.9398 |
| | | Pre. | 0.9531 | 0.9640 | 0.9587 | 0.9480 | | Pre. | 0.9510 | 0.9608 | 0.9570 | 0.9425 |
| | | Rec. | 0.9532 | 0.9639 | 0.9581 | 0.9444 | | Rec. | 0.9510 | 0.9607 | 0.9563 | 0.9398 |
| | | F1 | 0.9529 | 0.9636 | 0.9574 | 0.9434 | | F1 | 0.9507 | 0.9602 | 0.9555 | 0.9389 |
| FD | DFT | Acc. | 0.9320 | 0.9544 | 0.9497 | 0.9414 | FGA | Acc. | 0.9322 | 0.9506 | 0.9515 | 0.9293 |
| | | Pre. | 0.9321 | 0.9544 | 0.9497 | 0.9441 | | Pre. | 0.9321 | 0.9505 | 0.9518 | 0.9312 |
| | | Rec. | 0.9320 | 0.9544 | 0.9497 | 0.9414 | | Rec. | 0.9322 | 0.9506 | 0.9515 | 0.9293 |
| | | F1 | 0.9312 | 0.9538 | 0.9490 | 0.9404 | | F1 | 0.9316 | 0.9500 | 0.9508 | 0.9283 |
| NN | RCNN | Acc. | 0.9518 | 0.9629 | 0.9563 | 0.9376 | SCM | Acc. | 0.9490 | 0.9556 | 0.9497 | 0.9376 |
| | | Pre. | 0.9517 | 0.9629 | 0.9567 | 0.9410 | | Pre. | 0.9489 | 0.9555 | 0.9497 | 0.9410 |
| | | Rec. | 0.9518 | 0.9629 | 0.9563 | 0.9376 | | Rec. | 0.9490 | 0.9556 | 0.9497 | 0.9376 |
| | | F1 | 0.9516 | 0.9626 | 0.9556 | 0.9364 | | F1 | 0.9489 | 0.9552 | 0.9491 | 0.9364 |
| DL | CNNSP-SCM | Acc. | 0.9381 | 0.9629 | 0.9413 | 0.9368 | CNNSTFT-CC | Acc. | 0.9476 | 0.9556 | 0.9421 | 0.9293 |
| | | Pre. | 0.9379 | 0.9629 | 0.9414 | 0.9386 | | Pre. | 0.9475 | 0.9556 | 0.9420 | 0.9320 |
| | | Rec. | 0.9381 | 0.9629 | 0.9413 | 0.9368 | | Rec. | 0.9476 | 0.9556 | 0.9421 | 0.9293 |
| | | F1 | 0.9378 | 0.9626 | 0.9404 | 0.9360 | | F1 | 0.9474 | 0.9552 | 0.9412 | 0.9281 |
| | ENN-CC | Acc. | 0.9522 | 0.9599 | 0.9570 | 0.9429 | ENN-CI | Acc. | 0.9506 | 0.9629 | 0.9592 | 0.9451 |
| | | Pre. | 0.9521 | 0.9600 | 0.9575 | 0.9461 | | Pre. | 0.9505 | 0.9632 | 0.9599 | 0.9489 |
| | | Rec. | 0.9522 | 0.9599 | 0.9570 | 0.9429 | | Rec. | 0.9506 | 0.9629 | 0.9592 | 0.9451 |
| | | F1 | 0.9519 | 0.9595 | 0.9563 | 0.9419 | | F1 | 0.9504 | 0.9625 | 0.9585 | 0.9441 |
| | MAM-FGA | Acc. | 0.9333 | 0.9372 | 0.9479 | 0.9301 | MAM-PR | Acc. | 0.9298 | 0.9372 | 0.9461 | 0.9316 |
| | | Pre. | 0.9335 | 0.9369 | 0.9482 | 0.9335 | | Pre. | 0.9299 | 0.9368 | 0.9461 | 0.9345 |
| | | Rec. | 0.9333 | 0.9372 | 0.9479 | 0.9301 | | Rec. | 0.9298 | 0.9372 | 0.9461 | 0.9316 |
| | | F1 | 0.9326 | 0.9365 | 0.9471 | 0.9287 | | F1 | 0.9290 | 0.9365 | 0.9453 | 0.9304 |
| | MLP1STFT | Acc. | 0.9618 | 0.9656 | 0.9626 | 0.9388 | MLP1WT-SCM | Acc. | 0.9560 | 0.9647 | 0.9596 | 0.9444 |
| | | Pre. | 0.9619 | 0.9658 | 0.9635 | 0.9404 | | Pre. | 0.9559 | 0.9649 | 0.9603 | 0.9480 |
| | | Rec. | 0.9618 | 0.9656 | 0.9626 | 0.9388 | | Rec. | 0.9560 | 0.9647 | 0.9596 | 0.9444 |
| | | F1 | 0.9615 | 0.9652 | 0.9620 | 0.9380 | | F1 | 0.9557 | 0.9643 | 0.9589 | 0.9434 |
| ML | KNN | Acc. | 0.9480 | 0.9569 | 0.9547 | 0.9357 | BDT-PCNN | Acc. | 0.9472 | 0.9544 | 0.9512 | 0.9391 |
| | | Pre. | 0.9479 | 0.9568 | 0.9550 | 0.9387 | | Pre. | 0.9471 | 0.9543 | 0.9512 | 0.9424 |
| | | Rec. | 0.9480 | 0.9569 | 0.9547 | 0.9357 | | Rec. | 0.9472 | 0.9544 | 0.9512 | 0.9391 |
| | | F1 | 0.9478 | 0.9565 | 0.9541 | 0.9346 | | F1 | 0.9470 | 0.9539 | 0.9505 | 0.9380 |

signal count diminishes, increasing the statistical impact of each mislabeled event.

Several methods demonstrate consistently strong performance across different energy conditions. In DL, MLP1STFT stands out with high F1-scores of 0.9615 at 2 MeV, peaking at 0.9652 at 3 MeV, 0.9620 at 4 MeV, and 0.9380 at 5 MeV. Similarly, ENN-CI shows strong F1-scores ranging from 0.9441 at 5 MeV to 0.9625 at 3 MeV. In the TD, CI and CC show robust results, with CI's F1-scores ranging from 0.9434 to 0.9636 and CC from 0.9389 to 0.9602. Moreover, NN methods like RCNN also perform well, with F1-scores ranging from 0.9364 to 0.9626. In ML, methods such as KNN exhibit solid performance with F1-scores ranging from 0.9346 to 0.9565, while BDT-PCNN maintains scores from 0.9380 to 0.9539, demonstrating reliability across thresholds.

These trends highlight the robustness of certain PSD methods in handling varying energy levels, while others struggle more with lower energy signals. Figure 7 provides a radar chart illustrating the performance of selected methods under these varying energy levels. For each taxonomy, the two best, two middle, and two poorest performing methods are presented, offering a visual comparison of their strengths and weaknesses. Most TD and NN methods perform stably under various energy conditions. In contrast, mid-level FD and ML methods consistently underperform in PSD, demonstrating that only the top-performing methods in these two categories are robust enough. Although DL methods often perform outstandingly even among mid-range options, they are not proven to be generally robust across all variants. This is because numerous MLP-based methods in the DL category excel at PSD, essentially dominating the performance rankings under any threshold. However, many other models (e.g., CNN-based models) exhibit poor PSD performance. Details on the performance of all PSD methods are available in Supplementary Information Tables S2, S3, and S4.

## VIII. Discussion

The application of ML and DL to PSD has revealed a complex and often counter-intuitive performance landscape where simpler models frequently outperform state-of-the-art architectures. This discussion delves into the reasons for this phenomenon by analyzing the critical interplay between model design, data characteristics, and evaluation methods. We structure our analysis around a series of key questions that challenge conventional wisdom in the field, starting with an examination of hybrid models where DL regressors can surpass their statistical 'teachers.' We then re-evaluate the classic trade-offs between statistical and prior-knowledge paradigms in an era of powerful hardware. Subsequently, we dissect the architectural suitability of MLPs, CNNs, Transformers, and Mamba for PSD's unique data properties, followed by a critical look at evaluation metrics beyond the standard FOM. Through this integrated analysis, we aim to explain why architectural simplicity often triumphs in PSD and how hybrid approaches offer a promising path forward.

### A. Can a Student Surpass its Teacher?

Experimental results challenge a common assumption in machine learning: that a model cannot perform better than the data it learns from. We found that certain DL and ML regressors, when trained on features from statistical methods, can indeed surpass the performance of their "teachers." For example, an MLP1 model trained on PCNN factors achieved a higher F1-score (0.952) than PCNN alone (0.948), and an LSTM model dramatically improved upon the underperforming FS method, boosting the F1-score from 0.590 to 0.910.

This happens because DL models are not mere mimics; they can act as powerful "amplifiers". Their architectural advantages, like the ability to learn hierarchical representations or model temporal dependencies, allow them to refine and enhance the features provided by statistical methods. They can uncover subtle, non-linear patterns that the original statistical rules miss, effectively extrapolating beyond the teacher's explicit knowledge.

However, this amplification is not guaranteed. A model is still significantly influenced by its teacher, and a poor one can limit its student's potential by providing noisy or suboptimal features. While a gifted student can sometimes overcome a flawed teacher, selecting a robust statistical method as a starting point remains crucial for achieving top-tier performance and stability. This highlights the promise of hybrid models, which combine the best of both worlds, but also the importance of choosing the right foundation.

### B. Statistical or Prior-Knowledge PSD?

Statistical discrimination methods and prior-knowledge discrimination methods each bring distinct characteristics to PSD tasks, with traditional trade-offs in power, efficiency, and adaptability that are increasingly blurred by technological advancements. Statistical methods, encompassing TD (e.g., CC, CI), FD (e.g., DFT, FGA), and NN (e.g., PCNN, RCNN), rely on predefined mathematical formulations or biologically inspired rules to extract features like charge ratios or spectral components directly from pulse waveforms. These methods are typically interpretable, require no training data, and excel in computational efficiency, making them ideal for real-time applications with limited resources. However, they often lack flexibility in handling complex, noisy, or variable data, leading to performance degradation in challenging scenarios, as evidenced by lower F1-scores (e.g., FS at 0.590) and failed FOMs in our benchmarks.

In contrast, prior-knowledge methods, including ML (e.g., BDT, KNN) and DL (e.g., MLPs, RNNs), learn discriminative patterns from labeled data, enabling them to model intricate non-linear interactions and achieve higher accuracy in diverse conditions. These approaches are generally more powerful, as seen in top F1-scores like MLP2STFT's 0.962, but historically come at the cost of greater computational demands and reduced interpretability due to their "black-box" nature. Conventionally, the trade-off is clear: statistical methods offer efficiency and simplicity at the expense of raw discriminative power, while prior-knowledge methods provide superior performance but require more resources for training and inference.

However, as shown in Table 6's time consumption data, rapid GPU advancements have significantly narrowed this efficiency gap; many DL and ML models now execute in comparable or even shorter times than statistical methods, thanks to parallel computation capabilities. For example, optimized regressors like CNNSHAL complete inference faster than computationally intensive statistical approaches like CI, despite their complexity. Not to mention even simpler DL models such as ENN and MLP1 series, with top-tier performance and inference time at the scale of a few milliseconds (for processing about 5,000 pulse signals). This is faster than any statistical method when specific hardware optimization is not applied. Moreover, building on the previous discussion, DL and ML models can surpass their statistical "teachers" in hybrid setups, as regressors refine and enhance extracted features for better generalization. Thus, one need not strictly choose between these trade-offs. A possibly superior solution lies in hybrid models: employing an ML or DL regressor to learn feature extraction from a top-tier statistical method, preferably from neural network (e.g., PCNN, RCNN) or time-domain (e.g., CC and CI) categories.

TABLE VI. INFERENCE TIME CONSUMPTION FOR STATISTICAL AND PRIOR-KNOWLEDGE PSD METHOD

| Tax. | Method | | | | | | |
|---|---|---|---|---|---|---|---|
| Time-domain | CC | CI | FEPS | GP | LLR | LMT | PCA |
| | 0.030±0.000 | 0.050±0.011 | 0.036±0.005 | 0.010±0.000 | 0.014±0.005 | 0.028±0.004 | 0.184±0.014 |
| | PGA | PR | ZC | | | | |
| | 0.018±0.007 | 0.030±0.000 | 2.618±0.028 | | | | |
| Frequency-domain | FGA | SDCC | DFT | WT1 | WT2 | FS | SD |
| | 0.084±0.008 | 0.020±0.006 | 0.222±0.004 | 40.116±0.154 | 0.494±0.016 | 0.300±0.006 | 10.158±0.048 |
| Neural network | LG | PCNN | RCNN | SCM | | | |
| | 9.718±0.076 | 3.606±0.022 | 49.162±0.391 | 4.814±0.015 | | | |
| Deep learning | CNNDEEP-Clf | CNNDEEP-Reg | CNNFT-Clf | CNNFT-Reg | CNNSHAL-Clf | CNNSHAL-Reg | CNNSP-Clf |
| | 51.726±0.367 | 50.862±0.238 | 3.260±0.077 | 3.712±0.640 | 0.051±0.038 | 0.050±0.023 | 2.248±0.367 |
| | CNNSP-Reg | CNNSTFT-Clf | CNNSTFT-Reg | CNNWT-Clf | CNNWT-Reg | ENN-Clf | ENN-Reg |
| | 2.368±0.375 | 3.296±0.117 | 3.124±0.151 | 50.274±0.132 | 50.558±0.950 | 0.0084±0.0005 | 0.005±0.002 |
| | GRU-Clf | GRU-Reg | LSTM-Clf | LSTM-Reg | MAM-Clf | MAM-Reg | MLP1-Clf |
| | 0.0164±0.0053 | 0.013±0.001 | 0.0143±0.0027 | 0.014±0.005 | 0.482±0.072 | 0.568±0.088 | 0.0084±0.0005 |
| | MLP1-Reg | MLP1FT-Clf | MLP1FT-Reg | MLP1PCA-Clf | MLP1PCA-Reg | MLP1STFT-Clf | MLP1STFT-Reg |
| | 0.005±0.001 | 0.0091±0.0014 | 0.009±0.003 | 0.0071±0.0008 | 0.011±0.005 | 0.0081±0.0005 | 0.007±0.000 |
| | MLP1WT-Clf | MLP1WT-Reg | MLP2-Clf | MLP2-Reg | MLP2FT-Clf | MLP2FT-Reg | MLP2PCA-Clf |
| | 0.0043±0.0003 | 0.053±0.014 | 0.0151±0.0143 | 0.031±0.010 | 0.0106±0.0007 | 0.022±0.020 | 0.0078±0.0008 |
| | MLP2PCA-Reg | MLP2STFT-Clf | MLP2STFT-Reg | MLP2WT-Clf | MLP2WT-Reg | MLP3-Clf | MLP3-Reg |
| | 0.009±0.004 | 0.0075±0.0014 | 0.012±0.009 | 0.0394±0.0010 | 0.042±0.005 | 0.0072±0.0013 | 0.004±0.001 |
| | MLP3FT-Clf | MLP3FT-Reg | MLP3PCA-Clf | MLP3PCA-Reg | MLP3STFT-Clf | MLP3STFT-Reg | MLP3WT-Clf |
| | 0.0106±0.0005 | 0.007±0.001 | 0.0074±0.0016 | 0.004±0.001 | 0.0081±0.0007 | 0.005±0.001 | 0.0367±0.0011 |
| | MLP3WT-Reg | RNN-Clf | RNN-Reg | TRAN-Clf | TRAN-Reg | | |
| | 0.033±0.001 | 0.0124±0.0003 | 0.008±0.000 | 0.9147±0.0018 | 1.707±0.961 | | |
| Machine learing | BDT-Clf | BDT-Reg | DT-Clf | DT-Reg | FCM-Clf | GMM-Clf | KNN-Clf |
| | 0.1450±0.0077 | 0.139±0.004 | 0.0167±0.0037 | 0.008±0.004 | 0.0618±0.0038 | 0.0285±0.0043 | 0.1228±0.0062 |
| | KNN-Reg | LINRE-Reg | LOGRE-Clf | LRSTFT-Clf | LVQ-Clf | SVM-Clf | TEM-Clf |
| | 0.002±0.002 | 0.005±0.003 | 0.0114±0.0047 | 3.5237±0.0199 | 0.0087±0.0038 | 0.0382±0.0057 | 36.6644±0.0024 |

a. Inference time, measured in seconds, refers to the duration required to process approximately 5,000 signals.

This synergy harnesses GPU-accelerated computation, robust statistical engineering, and the generalization capability of prior-knowledge models, yielding systems that are both efficient and highly effective for PSD.

### C. Why Do MLPs Excel in PSD?

MLPs often demonstrate superior performance in PSD tasks because their simple, fully connected architecture is remarkably well-suited to the nature of pulse data. In PSD, the most telling information, such as the decay tail where neutron and gamma pulses diverge, is typically found in specific and consistent time segments of the waveform. Unlike more complex models that assume features might appear anywhere, MLPs assign independent weights to every time sample [73]. This allows them to directly learn the importance of these fixed locations without the computational overhead of searching for patterns [74]. Their ability to take a "global view" of the pulse lets them effectively memorize and exploit subtle, position-specific details that are key for discrimination. This direct approach makes MLPs highly effective and efficient, consistently achieving top-tier accuracy in both classification and regression tasks, often with less risk of being undertrained on the small datasets common in PSD.

Furthermore, MLPs are robust and versatile. They adapt

well whether they are processing raw pulse signals or features that have been pre-extracted by methods like STFT or WT. This flexibility, combined with their general resilience to variations in hyperparameters and model depth, makes them a practical and powerful choice. For PSD, where the specific shape of the pulse is more important than long-range context, the architectural simplicity of an MLP is a significant advantage.

### D. Why Do CNNs Struggle with PSD?

CNNs frequently underperform in PSD applications because their core strength, known as translation invariance [75], becomes a liability. CNNs are designed to find important patterns no matter where they appear in the data, which is ideal for tasks like image processing [76]. However, in PSD, the key discriminative features are in fixed temporal locations. A CNN's convolutional filters search the entire signal for local patterns, an unnecessary effort that can lead it astray. This architectural mismatch can cause the model to miss the key features entirely or to overfit by memorizing irrelevant noise, especially on smaller datasets. While CNNs may be adequate for some regression tasks, they often fall short in classification, where precise decision boundaries require a more holistic understanding of the pulse shape.

This problem is often magnified when researchers try to adapt CNNs by converting 1D pulse signals into 2D representations like spectrograms. While visually informative, this transformation can spread out the very localized, time-domain features that are critical for discrimination. This forces the CNN to work even harder to relearn connections that were explicit in the original waveform, increasing the risk of overfitting in data-scarce environments. Such transformations can inadvertently blur the essential details, leading to the variable and often disappointing performance of CNNs in PSD. Without significant modifications, standard CNNs are simply not a natural fit for the direct, position-specific analysis that PSD requires.

### E. Why Does Mamba Struggle with PSD?

Mamba models, while highly efficient for processing long data sequences, are generally not well-suited for PSD tasks. Their key innovation SSM introduces a layer of complexity that is unnecessary for the short and self-contained nature of pulse waveforms. This mechanism is designed to adaptively filter information over long periods, a powerful feature for tasks like language modeling but overkill for PSD pulses [77], which are typically less than a thousand samples long. For these short signals, only even shorter parts are relevant, and Mamba's selective gating adds computational overhead without a clear benefit. This can result in slower training and inference compared to simpler architectures like MLPs.

Moreover, Mamba's performance can be quite sensitive to its hyperparameter settings [78], such as state dimensions and convolution widths. Fine-tuning these is challenging with the small datasets typical in PSD, increasing the risk of overfitting. Its framework, optimized for capturing dynamic temporal patterns, may struggle to lock onto the static, shape-based features that define neutron and gamma pulses. This architectural misalignment makes Mamba a less robust choice for PSD, often requiring larger model structures or extensive fine-tuning to become competitive, which undermines the goal of creating efficient and reliable systems.

### F. Why Do Transformers Struggle with PSD?

Transformers often underperform in PSD because their central mechanism, self-attention, is poorly matched to the problem's characteristics. Self-attention excels at identifying complex, long-range dependencies by calculating relationships between all parts of a sequence [79]. While powerful, this approach is computationally expensive and inefficient for PSD, where discrimination relies on local features within a short signal, not on context from distant time points. For a typical pulse, the vast majority of these calculations are redundant, adding significant computational load without improving accuracy.

This architectural mismatch creates practical challenges. The high data and tuning requirements [80] of Transformers make them difficult to implement effectively in the data-limited environments common to PSD. Without enough data or careful tuning, they can easily overfit to noise or fail to learn the simple, shape-based patterns that MLPs capture so readily. Even with positional encodings designed to provide location information, a Transformer's fundamental bias toward permutation-invariant, global relationships can obscure the very local morphological details that matter most. This makes Transformers a less efficient and less reliable choice for PSD compared to simpler models that are better aligned with the task's local nature.

### G. Evaluating PSD Beyond FOM

While the FOM is the traditional benchmark for PSD performance, a more complete assessment requires looking at a broader set of metrics, especially when ground-truth labels are available. The F1-score, which provides a balanced measure of precision and recall, offers a direct evaluation of a model's classification accuracy. Another powerful tool is the ROC curve and its corresponding AUC value, which illustrates a model's ability to discriminate between classes across all possible thresholds. A high AUC value, approaching 1.0, signals robust and reliable performance, making it an excellent way to compare different model architectures on a level playing field.

In practical scenarios where ground-truth labels may not exist, correlation-based analysis provides a valuable alternative. By calculating the Pearson correlation coefficients between the outputs of different PSD methods, we can gauge their consistency. High-performing methods tend to show strong correlations with one another, suggesting they are identifying the same underlying physical features. In contrast, weak correlations often indicate that a method is diverging into noise. Together, these alternative metrics, F1-score, ROC/AUC, and correlation analysis, overcome some of the limitations of FOM, such as its sensitivity to histogram binning. This multi-faceted approach to evaluation provides a more nuanced and reliable understanding of a model's true performance.

## IX. CONCLUSION

In this work, we have presented the most extensive comparative study of Pulse Shape Discrimination (PSD) algorithms to date. By implementing and benchmarking nearly sixty distinct methods across both statistical and prior-knowledge paradigms on two unified experimental datasets, our research provides a clear and comprehensive overview of the current state of the field. The results unequivocally demonstrate the superior performance of deep learning approaches, particularly Multi-Layer Perceptrons (MLPs) and hybrid models that pair statistical feature extraction with neural network regression. These advanced methods consistently achieve high F1-scores, often exceeding 0.95, and significantly outperform traditional statistical baselines.

Our analysis offers several key insights into the architectural considerations for PSD. We have highlighted the distinct advantages of MLPs, whose structure is well-suited for capturing position-specific features critical for discrimination, while also explaining the architectural mismatches that cause CNNs, Transformers, and Mamba models to struggle with the short-sequence nature of pulse data. Furthermore, we have addressed the inherent limitations of conventional evaluation metrics like the Figure of Merit (FOM), advocating instead for a more robust and multifaceted assessment strategy that incorporates the F1-score, ROC-AUC, and inter-method correlations. The consistent and robust performance of the top-performing methods across a range of energy thresholds further underscores their practical applicability for real-world radiation detection scenarios.

To further support the research community, we have made our comprehensive, dual-language (Python and MATLAB) toolbox and the standardized datasets used in this study open access. It is our hope that this resource, available at https://github.com/HaoranLiu507/PulseShapeDiscrimination, will be instrumental in facilitating reproducibility, fostering further innovation, and accelerating advancements in PSD for critical applications in nuclear physics, radiation monitoring, and beyond.

Future research should explore several promising avenues to advance PSD capabilities. One key direction is the development of novel hybrid architectures that can more effectively fuse statistical features with deep learning models. Another is the expansion of training corpora to include larger and more diverse datasets, which is critical for improving model generalization. Furthermore, future work should investigate the integration of multi-modal data into DL frameworks; beyond the pulse signal itself, incorporating contextual information such as the scintillator type, neutron source, and environmental conditions like temperature could significantly enhance discrimination accuracy. Finally, a continued focus on optimizing models for real-time, on-the-fly implementation will be crucial for transitioning these advanced techniques from research to practical field applications.

ACKNOWLEDGMENT

The authors thank Nicholai Mauritzson and Kevin G Fissum from Division of Nuclear Physics, Lund University, for invaluable assistance regarding time-of-flight methodology.

REFERENCES

[1] M. L. Roush, M. A. Wilson, and W. F. Hornyak, "Pulse shape discrimination," *Nucl. Instrum. Methods,* vol. 31, no. 1, pp. 112–124, 1964, doi: 10.1016/0029-554X(64)90333-7.

[2] T. Bily and L. Keltnerova, "Non-linearity assessment of neutron detection systems using zero-power reactor transients," *Appl. Radiat. Isot.,* vol. 157, p. 109016, 2020, doi: 10.1016/j.apradiso.2019.109016.

[3] E. Rohée *et al.*, "Delayed Neutron Detection with graphite moderator for clad failure detection in Sodium-Cooled Fast Reactors," *Ann. Nucl. Energy,* vol. 92, pp. 440–446, 2016, doi: 10.1016/j.anucene.2016.02.003.

[4] Y. Kavun, T. Eyyup, M. Şahan, and A. Salan, "Calculation of Production Reaction Cross Section of Some Radiopharmaceuticals Used in Nuclear Medicine by New Density Dependent Parameters," *Süleyman Demirel Üniversitesi Fen Edebiyat Fakültesi Fen Dergisi,* vol. 14, no. 1, pp. 57–61, 2019, doi: 10.29233/sdufeffd.477539.

[5] D. VanDerwerken *et al.*, "Meteorologically Driven Neutron Background Prediction for Homeland Security," *IEEE Trans. Nucl. Sci.,* vol. 65, no. 5, pp. 1187–1195, 2018, doi: 10.1109/TNS.2018.2821630.

[6] A. Glenn, Q. Cheng, A. D. Kaplan, and R. Wurtz, "Pulse pileup rejection methods using a two-component Gaussian Mixture Model for fast neutron detection with pulse shape discriminating scintillator," *Nucl. Instrum. Methods Phys. Res., Sect. A,* vol. 988, p. 164905, 2021, doi: 10.1016/j.nima.2020.164905.

[7] F. C. E. Teh *et al.*, "Value-Assigned Pulse Shape Discrimination for Neutron Detectors," *IEEE Trans. Nucl. Sci.,* vol. 68, no. 8, pp. 2294–2300, 2021, doi: 10.1109/TNS.2021.3091126.

[8] R. L. Garnett and S. H. Byun, "NeutralNet: Development and testing of a machine learning solution for pulse shape discrimination," *Appl. Radiat. Isot.,* vol. 211, p. 111384, 2024, doi: 10.1016/j.apradiso.2024.111384.

[9] H. Liu, P. Li, M. Liu, K. Wang, Z. Zuo, and B. Liu, "Pulse Shape Discrimination Based on the Tempotron: A Powerful Classifier on GPU," *IEEE Trans. Nucl. Sci.,* vol. 71, no. 10, pp. 2297–2308, 2024, doi: 10.1109/TNS.2024.3444888.

[10] M. Moszynski *et al.*, "Study of n-γ discrimination by digital charge comparison method for a large volume liquid scintillator," *Nucl. Instrum. Methods Phys. Res., Sect. A,* vol. 317, no. 1, pp. 262–272, 1992, doi: 10.1016/0168-9002(92)90617-D.

[11] E. Gatti and F. D. Martini, "A new linear method of discrimination between elementary particles in scintillation counters," in *Nuclear Electronics II. Proceedings of the Conference on Nuclear Electronics. V. II*, 1962, pp. 265–276.

[12] B. D'Mellow, M. D. Aspinall, R. O. Mackin, M. J. Joyce, and A. J. Peyton, "Digital discrimination of neutrons and γ-rays in liquid scintillators using pulse gradient analysis," *Nucl. Instrum. Methods Phys. Res., Sect. A,* vol. 578, no. 1, pp. 191–197, 2007, doi: 10.1016/j.nima.2007.04.174.

[13] G. Liu, M. J. Joyce, X. Ma, and M. D. Aspinall, "A digital method for the discrimination of neutrons and gamma rays with organic scintillation detectors using frequency gradient analysis," *IEEE Trans. Nucl. Sci.,* vol. 57, no. 3, pp. 1682–1691, 2010, doi: 10.1109/TNS.2010.2044246.

[14] M. J. Safari, F. A. Davani, H. Afarideh, S. Jamili, and E. Bayat, "Discrete Fourier Transform Method for Discrimination of Digital Scintillation Pulses in Mixed Neutron-Gamma Fields," *IEEE Trans. Nucl. Sci.,* vol. 63, no. 1, pp. 325–332, 2016, doi: 10.1109/TNS.2016.2514400.

[15] S. Yousefi, L. Lucchese, and M. D. Aspinall, "Digital discrimination of neutrons and gamma-rays in liquid scintillators using wavelets," *Nucl. Instrum. Methods Phys. Res., Sect. A,* vol. 598, no. 2, pp. 551–555, 2009, doi: 10.1016/j.nima.2008.09.028.

[16] H. Liu, Y. Cheng, Z. Zuo, T. Sun, and K. Wang, "Discrimination of neutrons and gamma rays in plastic scintillator based on pulse-coupled neural network," *Nucl. Sci. Tech.,* vol. 32, no. 8, p. 82, 2021, doi: 10.1007/s41365-021-00915-w.

[17] H. Liu, M. Liu, Y. Xiao, P. Li, Z. Zuo, and Y. Zhan, "Discrimination of neutron and gamma ray using the ladder gradient method and analysis of filter adaptability," *Nucl. Sci.*

[18] B.-Q. Liu, H.-R. Liu, L. Chang, Y.-X. Cheng, Z. Zuo, and P. Li, "Discrimination of neutrons and gamma-rays in plastic scintillator based on spiking cortical model," *Nucl. Eng. Technol.,* vol. 55, no. 9, pp. 3359–3366, 2023, doi: 10.1016/j.net.2023.04.032.

[19] T. S. Sanderson, C. D. Scott, M. Flaska, J. K. Polack, and S. A. Pozzi, "Machine learning for digital pulse shape discrimination," in *2012 IEEE Nuclear Science Symposium and Medical Imaging Conference Record (NSS/MIC)*, 27 Oct.–3 Nov. 2012 2012, pp. 199–202, doi: 10.1109/NSSMIC.2012.6551092.

[20] M. Durbin, M. A. Wonders, M. Flaska, and A. T. Lintereur, "K-Nearest Neighbors regression for the discrimination of gamma rays and neutrons in organic scintillators," *Nucl. Instrum. Methods Phys. Res., Sect. A,* vol. 987, p. 164826, 2021, doi: 10.1016/j.nima.2020.164826.

[21] E. Yun, J. Y. Choi, S. Y. Kim, and K. K. Joo, "Pulse Shape Discrimination of n/γ in Liquid Scintillator at PMT Nonlinear Region Using Artificial Neural Network Technique," *Sensors,* vol. 24, no. 24, p. 8060, 2024, doi: 10.3390/s24248060.

[22] J. Griffiths, S. Kleinegesse, D. Saunders, R. Taylor, and A. Vacheret, "Pulse shape discrimination and exploration of scintillation signals using convolutional neural networks," *Mach. Learn.: Sci. Technol.,* vol. 1, no. 4, p. 045022, 2020, doi: 10.1088/2632-2153/abb781.

[23] A. Vaswani *et al.*, "Attention is all you need," in *Proceedings of the 31st International Conference on Neural Information Processing Systems*, Long Beach, California, USA, 2017: Curran Associates Inc., pp. 6000–6010, doi: 10.5555/3295222.3295349.

[24] G. H. V. Bertrand, M. Hamel, S. Normand, and F. Sguerra, "Pulse shape discrimination between (fast or thermal) neutrons and gamma rays with plastic scintillators: State of the art," *Nucl. Instrum. Methods Phys. Res., Sect. A,* vol. 776, pp. 114–128, 2015, doi: 10.1016/j.nima.2014.12.024.

[25] N. P. Zaitseva *et al.*, "Recent developments in plastic scintillators with pulse shape discrimination," *Nucl. Instrum. Methods Phys. Res., Sect. A,* vol. 889, pp. 97–104, 2018, doi: 10.1016/j.nima.2018.01.093.

[26] A. Foster *et al.*, "On the fabrication and characterization of heterogeneous composite neutron detectors with triple-pulse-shape-discrimination capability," *Nucl. Instrum. Methods Phys. Res., Sect. A,* vol. 954, p. 161681, 2020, doi: 10.1016/j.nima.2018.11.140.

[27] I. Ahnouz, H. Arahmane, and R. Sebihi, "A Review of Neutron–Gamma-Ray Discrimination Methods Using Organic Scintillators," *Nucl. Sci. Eng.,* vol. 198, no. 12, pp. 2241–2273, 2024, doi: 10.1080/00295639.2024.2316946.

[28] I. A. Pawełczak, S. A. Ouedraogo, A. M. Glenn, R. E. Wurtz, and L. F. Nakae, "Studies of neutron–γ pulse shape discrimination in EJ-309 liquid scintillator using charge integration method," *Nucl. Instrum. Methods Phys. Res., Sect. A,* vol. 711, pp. 21–26, 2013, doi: 10.1016/j.nima.2013.01.028.

[29] Z. Zuo, Y. Xiao, Z. Liu, B. Liu, and Y. Yan, "Discrimination of neutrons and gamma-rays in plastic scintillator based on falling-edge percentage slope method," *Nucl. Instrum. Methods Phys. Res., Sect. A,* vol. 1010, p. 165483, 2021, doi: 10.1016/j.nima.2021.165483.

[30] M. Akashi-Ronquest *et al.*, "Improving photoelectron counting and particle identification in scintillation detectors with Bayesian techniques," *Astropart. Phys.,* vol. 65, pp. 40–54, 2015, doi: 10.1016/j.astropartphys.2014.12.006.

[31] P. Adhikari *et al.*, "Pulse-shape discrimination against low-energy Ar-39 beta decays in liquid argon with 4.5 tonne-years of DEAP-3600 data," *The European Physical Journal C,* vol. 81, no. 9, p. 823, 2021, doi: 10.1140/epjc/s10052-021-09514-w.

[32] H. S. Lee *et al.*, "Neutron calibration facility with an Am-Be source for pulse shape discrimination measurement of CsI(Tl) crystals," *J. Instrum.,* vol. 9, no. 11, p. P11015, 2014, doi: 10.1088/1748-0221/9/11/P11015.

[33] T. Alharbi, "Principal Component Analysis for pulse-shape discrimination of scintillation radiation detectors," *Nucl. Instrum. Methods Phys. Res., Sect. A,* vol. 806, pp. 240–243, 2016, doi: 10.1016/j.nima.2015.10.030.

[34] D. Takaku, T. Oishi, and M. Baba, "Development of neutron-gamma discrimination technique using pattern-recognition method with digital signal processing," *Prog. Nucl. Sci. Technol,* vol. 1, pp. 210–213, 2011, doi: 10.15669/pnst.1.210.

[35] P. Sperr, H. Spieler, M. R. Maier, and D. Evers, "A simple pulse-shape discrimination circuit," *Nucl. Instrum. Methods,* vol. 116, no. 1, pp. 55–59, 1974, doi: 10.1016/0029-554X(74)90578-3.

[36] S. Pai, W. F. Piel, D. B. Fossan, and M. R. Maier, "A versatile electronic pulse-shape discriminator," *Nucl. Instrum. Methods Phys. Res., Sect. A,* vol. 278, no. 3, pp. 749–754, 1989, doi: 10.1016/0168-9002(89)91199-6.

[37] M. Liu, B. Liu, Z. Zuo, L. Wang, G. Zan, and X. Tuo, "Toward a fractal spectrum approach for neutron and gamma pulse shape discrimination," *Chin. Phys. C,* vol. 40, no. 6, p. 066201, 2016, doi: 10.1088/1674-1137/40/6/066201.

[38] A. Abdelhakim and E. Elshazly, "Efficient pulse shape discrimination using scalogram image masking and decision tree," *Nucl. Instrum. Methods Phys. Res., Sect. A,* vol. 1050, p. 168140, 2023, doi: 10.1016/j.nima.2023.168140.

[39] D. I. Shippen, M. J. Joyce, and M. D. Aspinall, "A Wavelet Packet Transform Inspired Method of Neutron-Gamma Discrimination," *IEEE Trans. Nucl. Sci.,* vol. 57, no. 5, pp. 2617–2624, 2010, doi: 10.1109/TNS.2010.2044190.

[40] W. G. J. Langeveld, M. J. King, J. Kwong, and D. T. Wakeford, "Pulse Shape Discrimination Algorithms, Figures of Merit, and Gamma-Rejection for Liquid and Solid Scintillators," *IEEE Trans. Nucl. Sci.,* vol. 64, no. 7, pp. 1801–1809, 2017, doi: 10.1109/TNS.2017.2681654.

[41] H. Liu, Z. Zuo, P. Li, B. Liu, L. Chang, and Y. Yan, "Anti-noise performance of the pulse coupled neural network applied in discrimination of neutron and gamma-ray," *Nucl. Sci. Tech.,* vol. 33, no. 6, p. 75, 2022, doi: 10.1007/s41365-022-01054-6.

[42] H. Liu *et al.*, "Random-Coupled Neural Network," *Electronics,* vol. 13, no. 21, p. 4297, 2024, doi: 10.3390/electronics13214297.

[43] Y. LeCun, Y. Bengio, and G. Hinton, "Deep learning," *Nature,* vol. 521, no. 7553, pp. 436–444, 2015, doi: 10.1038/nature14539.

[44] Z. Li, F. Liu, W. Yang, S. Peng, and J. Zhou, "A Survey of Convolutional Neural Networks: Analysis, Applications, and Prospects," *IEEE Trans. Neural Networks Learn. Syst.,* vol. 33, no. 12, pp. 6999–7019, 2022, doi: 10.1109/TNNLS.2021.3084827.

[45] A. Karmakar, A. Pal, G. Anil Kumar, Bhavika, Vivek, and M. Tyagi, "Neutron-gamma pulse shape discrimination for organic scintillation detector using 2D CNN based image classification," *Appl. Radiat. Isot.,* vol. 217, p. 111653, 2025, doi: 10.1016/j.apradiso.2024.111653.

[46] M. Riedmiller, "Advanced supervised learning in multi-layer perceptrons — From backpropagation to adaptive learning algorithms," *Comput. Stand. Interfaces,* vol. 16, no. 3, pp. 265–278, 1994, doi: 10.1016/0920-5489(94)90017-5.

[47] S. Dutta, S. Ghosh, S. Bhattacharya, and S. Saha, "Pulse shape simulation and discrimination using machine learning techniques," *J. Instrum.,* vol. 18, no. 03, p. P03038, 2023, doi: 10.1088/1748-0221/18/03/P03038.

[48] Y. Yu, X. Si, C. Hu, and J. Zhang, "A Review of Recurrent Neural Networks: LSTM Cells and Network Architectures," *Neural Comput.,* vol. 31, no. 7, pp. 1235–1270, 2019, doi: 10.1162/neco_a_01199.

[49] C.-X. Zhang *et al.*, "Discrimination of neutrons and γ-rays in liquid scintillator based on Elman neural network*," *Chin. Phys. C,* vol. 40, no. 8, p. 086204, 2016, doi: 10.1088/1674-1137/40/8/086204.

[50] A. Gu and T. Dao, "Mamba: Linear-time sequence modeling with selective state spaces," *arXiv,* 2023, doi: 10.48550/arXiv.2312.00752.

[51] M. I. Jordan and T. M. Mitchell, "Machine learning: Trends, perspectives, and prospects," *Science,* vol. 349, no. 6245, pp. 255–260, 2015, doi: 10.1126/science.aaa8415.

[52] D. Wolski, M. Moszyński, T. Ludziejewski, A. Johnson, W. Klamra, and Ö. Skeppstedt, "Comparison of n-γ discrimination by zero-crossing and digital charge comparison methods," *Nucl. Instrum. Methods Phys. Res., Sect. A,* vol. 360, no. 3, pp. 584–592, 1995, doi: 10.1016/0168-9002(95)00037-2.

[53] X. Luo, G. Liu, and J. Yang, "Neutron/Gamma Discrimination Utilizing Fuzzy C-Means Clustering of the Signal from the Liquid Scintillator," in *2010 First International Conference on Pervasive Computing, Signal Processing and Applications*, 17–19 Sept. 2010 2010, pp. 994–997, doi: 10.1109/PCSPA.2010.245.
[54] M. Gelfusa *et al.*, "Advanced pulse shape discrimination via machine learning for applications in thermonuclear fusion," *Nucl. Instrum. Methods Phys. Res., Sect. A,* vol. 974, p. 164198, 2020, doi: 10.1016/j.nima.2020.164198.
[55] A. Abdelhakim and E. Elshazly, "Neutron/gamma pulse shape discrimination using short-time frequency transform," *Analog Integr. Circuits Signal Process.,* vol. 111, no. 3, pp. 387–402, 2022, doi: 10.1007/s10470-022-02009-y.
[56] T. Tambouratzis, D. Chernikova, and I. Pzsit, "Pulse shape discrimination of neutrons and gamma rays using Kohonen artificial neural networks," *J. Artif. Intell. Soft Comput. Res.,* vol. 3, 2013, doi: 10.2478/jaiscr-2014-0006.
[57] K. Wang, H. Liu, P. Li, M. Liu, and Z. Zuo, "Dataset for neutron and gamma-ray pulse shape discrimination," *arXiv,* 2023, doi: 10.48550/arXiv.2305.18242.
[58] M. Moszyński, M. Kapusta, D. Wolski, W. Klamra, and B. Cederwall, "Properties of the YAP : Ce scintillator," *Nucl. Instrum. Methods Phys. Res., Sect. A,* vol. 404, no. 1, pp. 157–165, 1998, doi: 10.1016/S0168-9002(97)01115-7.
[59] J. R. M. Annand, B. E. Andersson, I. Akkurt, and B. Nilsson, "An NE213A TOF spectrometer for high resolution (γ,n) reaction measurements," *Nucl. Instrum. Methods Phys. Res., Sect. A,* vol. 400, no. 2, pp. 344–355, 1997, doi: 10.1016/S0168-9002(97)01021-8.
[60] N. Mauritzson *et al.*, "Technique for the measurement of intrinsic pulse-shape discrimination for organic scintillators using tagged neutrons," *Nucl. Instrum. Methods Phys. Res., Sect. A,* vol. 1039, p. 167141, 2022, doi: 10.1016/j.nima.2022.167141.
[61] J. Scherzinger *et al.*, "Tagging fast neutrons from an 241Am/9Be source," *Appl. Radiat. Isot.,* vol. 98, pp. 74–79, 2015, doi: 10.1016/j.apradiso.2015.01.003.
[62] A. V. Oppenheim, *Discrete-time signal processing*. Pearson Education India, 1999.
[63] T. W. Parks and C. S. Burrus, *Digital filter design*. Wiley-Interscience, 1987.
[64] A. Andreas, "Digital signal processing: Signals, systems, and filters," *London, San Juan, Toromti: McGrall-Hill Companies,* vol. 937, 2006.
[65] E. O. Brigham, *The fast Fourier transform and its applications*. Prentice-Hall, Inc., 1988.
[66] P. S. Diniz, *Adaptive filtering*. Springer, 1997.
[67] G. R. Arce, *Nonlinear signal processing: a statistical approach*. John Wiley & Sons, 2004.
[68] P. Maragos and R. Schafer, "Morphological filters--Part II: Their relations to median, order-statistic, and stack filters," *IEEE Trans. Acoust. Speech Signal Process.,* vol. 35, no. 8, pp. 1170–1184, 1987, doi: 10.1109/TASSP.1987.1165254.
[69] S. Smith, *Digital signal processing: a practical guide for engineers and scientists*. Newnes, 2003.
[70] D. L. Donoho and I. M. Johnstone, "Adapting to Unknown Smoothness via Wavelet Shrinkage," *J. Am. Stat. Assoc.,* vol. 90, no. 432, pp. 1200–1224, 1995, doi: 10.1080/01621459.1995.10476626.
[71] J. S. Lim, *Two-dimensional signal and image processing*. Prentice-Hall, Inc., 1990.
[72] F. J. Harris, "On the use of windows for harmonic analysis with the discrete Fourier transform," *Proc. IEEE,* vol. 66, no. 1, pp. 51–83, 1978, doi: 10.1109/PROC.1978.10837.
[73] D. E. Rumelhart, G. E. Hinton, and R. J. Williams, "Learning representations by back-propagating errors," *Nature,* vol. 323, no. 6088, pp. 533–536, 1986, doi: 10.1038/323533a0.
[74] K. Hornik, M. Stinchcombe, and H. White, "Multilayer feedforward networks are universal approximators," *Neural Netw.,* vol. 2, no. 5, pp. 359–366, 1989, doi: 10.1016/0893-6080(89)90020-8.
[75] Y. Lecun, L. Bottou, Y. Bengio, and P. Haffner, "Gradient-based learning applied to document recognition," *Proc. IEEE,* vol. 86, no. 11, pp. 2278–2324, 1998, doi: 10.1109/5.726791.
[76] A. Krizhevsky, I. Sutskever, and G. E. Hinton, "Imagenet classification with deep convolutional neural networks," *Adv. Neural Inf. Process. Syst.,* vol. 25, 2012, doi: 10.5555/2999134.2999257.
[77] B. N. Patro and V. S. Agneeswaran, "Mamba-360: Survey of state space models as transformer alternative for long sequence modelling: Methods, Applications, and Challenges," *Eng. Appl. Artif. Intell.,* vol. 159, p. 111279, 2025, doi: 10.1016/j.engappai.2025.111279.
[78] Y. Deng, Y. Zhang, and J. Zhang, "The Influence of Mamba Model Hyperparameters on Training Time and Accuracy," in *2024 IEEE International Conference on Smart Internet of Things (SmartIoT)*, 14–16 Nov. 2024 2024, pp. 293–300, doi: 10.1109/SmartIoT62235.2024.00052.
[79] S. Khan, M. Naseer, M. Hayat, S. W. Zamir, F. S. Khan, and M. Shah, "Transformers in Vision: A Survey," *ACM Comput. Surv.,* vol. 54, no. 10s, p. Article 200, 2022, doi: 10.1145/3505244.
[80] A. Wiemerslage, K. Gorman, and K. von der Wense, "Quantifying the Hyperparameter Sensitivity of Neural Networks for Character-level Sequence-to-Sequence Tasks," St. Julian's, Malta, March 2024: Association for Computational Linguistics, in Proceedings of the 18th Conference of the European Chapter of the Association for Computational Linguistics (Volume 1: Long Papers), pp. 674–689, doi: 10.18653/v1/2024.eacl-long.40.